\documentclass[runningheads]{llncs}
\usepackage[T1]{fontenc}
\usepackage{graphicx}
\usepackage{booktabs}
\usepackage[misc]{ifsym}
\newcommand{\corr}{(\Letter)}
\usepackage{mwe}

\usepackage[utf8]{inputenc} 
\usepackage{url}            
\usepackage{amsfonts}       
\usepackage{nicefrac}       
\usepackage{microtype}      
\usepackage{xcolor}         
\graphicspath{{plots/}}
\usepackage{subcaption}
\usepackage{comment}

\usepackage{geometry}
\usepackage{hyperref}
\hypersetup{colorlinks=true,citecolor=green,
filecolor=black,linkcolor=red,urlcolor=magenta,	
linktocpage,linktoc=page}

\makeatletter
\newcommand*\bigcdot{\mathpalette\bigcdot@{1.2}}
\newcommand*\bigcdot@[2]{\mathbin{\vcenter{\hbox{\scalebox{#2}{$\m@th#1\bullet$}}}}}
\makeatother

\usepackage{amsmath}
\usepackage{amssymb}
\usepackage{bm}
\DeclareMathOperator*{\argmax}{arg\,max}

\usepackage{float}
\usepackage{caption}
\usepackage{subcaption}
\usepackage{comment}

\usepackage[numbers, sort&compress]{natbib}
\begin{document}

\title{Dropout Regularization in Extended Generalized Linear Models based on Double Exponential Families}

\titlerunning{Dropout in Extended Generalized Linear Models}

\author{Benedikt Lütke Schwienhorst\inst{1} \corr \and
Lucas Kock\inst{2} \and
Nadja Klein\inst{3}\and
David J. Nott\inst{2}}

\authorrunning{Lütke Schwienhorst et al.}

\institute{Department of Mathematics, University of Hamburg, Germany \email{benedikt.luetke.schwienhorst@uni-hamburg.de} \and
Department of Statistics and Data Science, National
University of Singapore, Singapore \and
Research Center Trustworthy Data Science and Security and Department of Statistics,  Technische Universität Dortmund, Germany}

\toctitle{Dropout Regularization in Extended Generalized Linear Models based on Double Exponential Families}

\tocauthor{Benedikt~Lütke~Schwienhorst, Lucas~Kock, Nadja~Klein, David~J.~Nott}

\maketitle

\setcounter{footnote}{0}

\begin{abstract}
Even though dropout is a popular regularization technique, its theoretical properties are not fully understood. In this paper we study dropout regularization in extended generalized linear models based on double exponential families, for which the dispersion parameter can vary with the features. A theoretical analysis shows that dropout regularization prefers rare but important features in both the mean and dispersion, generalizing an earlier result for conventional generalized linear models. To illustrate, we apply dropout to adaptive smoothing with B-splines, where both the mean and dispersion parameters are modeled flexibly. The important B-spline basis functions can be thought of as rare features, and we confirm in experiments that dropout is an effective form of regularization for mean and dispersion parameters that improves on a penalized maximum likelihood approach with an explicit smoothness penalty. An application to traffic detection data from Berlin further illustrates the benefits of our method.

\keywords{B-splines \and double exponential families \and dropout regularization \and generalized linear models \and nonparametric estimation \and overdispersion and underdispersion.}
\end{abstract}

\section{Introduction}

Dropout regularization \cite{HinSriNit2012,Srivastava+hkss14} was introduced in the context of neural networks and has been successfully implemented in a large number of applications \citep{VanKotWojWar2014, MerKesNitSoc2017, TanLe2019}. In its original formulation, dropout omits a randomly chosen subset of features during each iteration of stochastic gradient descent (SGD) optimization of the training loss. It has been generalized in various ways, and is an example of a broader class of methods using randomly corrupted features in model training \citep[e.g.,~][]{BurSch1996,MaaCheTzrWei2013}.

This work considers dropout for extended generalized linear models (GLMs) based on double exponential families (DEFs). The DEF was introduced by \citet{Efr1986} and generalizes the natural exponential family (EF) by incorporating an additional dispersion parameter. \citet{Efr1986} also developed regression models where the distribution of the output follows a DEF, and both the the mean and
dispersion can vary with the features.  The extended GLMs of \citet{Efr1986} are related to extended quasi-likelihood methods \citep{LeeNel2000} and are particularly useful for count data.  If count data are modelled using a binomial or Poisson distribution, then there is no separate scale parameter, and the variance is a function of the mean. For real count data, the variance can be more or less than expected based on a binomial or Poisson mean-variance relationship. If the variance is larger than expected, this is referred to as overdispersion \citep{McCNel1998} and not taking it into account can result in unreliable inference \citep{DunSmy2018}. Underdispersion, where the variance is less than expected, can also occur, but is less common. \citet{McCNel1998} state that it should be assumed that overdispersion is present unless proven otherwise. Extended GLMs for double binomial or double Poisson  distributions are suitable alternatives to standard GLMs for count data exhibiting over- or underdispersion. We use dropout regularization in the mean and the dispersion to avoid overfitting and to prevent co-adaptation of features \citep{HelLon2015}.  

Overdispersed or underdispersed models for count data arise in important machine learning applications. One example is the use of Dirichlet-multinomial models, which generalize beta-binomial models, in topic modelling \citep[e.g.,~][]{BleNgJor2003}. Flexible models for count data also arise in large-scale regression applications \citep[e.g.,~][]{HanBlePow2011} and in mixture-of-experts models with GLM components. The subtleties that arise in the behaviour of dropout for DEF-GLMs will potentially arise in other overdispersed models, where the effect of regularization on the mean and variance needs to be considered jointly.

In conventional GLMs with canonical link functions, \citet{WagWanLia2013} have shown that dropout performs regularization which is first-order equivalent to L2 regularization, with a penalty matrix related to the empirical Fisher information. The form of the penalty favors rare but important features. In the neural network literature, previous work by \citet{Bis1995} has shown that adding Gaussian noise to the training data is equivalent to L2 regularization (also known as Tikhonov regularization). \citet{BalSad2013} analyze dropout for both linear and non-linear networks, obtaining related results.  \citet{WanMan2013} consider fast dropout based on analytically marginalizing the dropout noise. \citet{WeiKakTen2020} consider deep network architectures and identify both explicit and implicit regularization effects of dropout training, where the explicit effect of regularization is approximated by an L2 penalty term. Despite the connections between L2 regularization and dropout, dropout can behave very differently from both L1 and L2 regularization in some circumstances \citep{HelLon2015}. 

Our work generalizes the study of \citet{WagWanLia2013} to extended GLMs based on DEFs.  Our first contribution is to give a theoretical analysis of the behaviour of dropout for extended GLMs.  We discuss the way that the regularization parameters for the mean and dispersion models interact, and the effect of over- or underdispersion on the regularization of the mean. We illustrate that dropout regularization in DEF-GLMs can be regarded as a form of L2 regularization, where the penalty matrix depends on the Fisher information for the mean and dispersion parameters. Understanding of the induced penalty requires considering penalization of the mean and dispersion parameters jointly, and penalization of the dispersion model behaves differently to penalization in the mean, with an asymmetry in the treatment of over- and underdispersion. Our paper contributes to the understanding of dropout regularization beyond its traditional application in neural networks, demonstrating its applicability in other model classes. Our second contribution is to consider the use of dropout regularization in nonparametric estimation of extended GLMs with B-splines, and to compare dropout with penalized maximum likelihood estimation (PMLE) \citep{GijProCla2010} in this setting. \citet{GijProCla2010} consider an explicit smoothness penalty based on second-order difference operators for regularization \citep{EilMar1996}. In accordance with our theoretical analysis, our experiments demonstrate that dropout regularization can be particularly effective when important B-spline basis functions are analogous to rare features in the mean and dispersion model. We also verify that the performance improves when there is only overdispersion and no underdispersion. After having confirmed the theory in our simulations, we illustrate the efficacy of dropout for DEF-GLMs with a real data set on traffic detection in Berlin.

The rest of this paper is organized as follows. Section~\ref{sec:prelim} introduces the extended GLM based on DEFs and Section~\ref{sec:dropout_mean_disp} analyzes the regularization induced by dropout in this model class. Then, an application to adaptive smoothing with B-splines based on simulated and real world data is discussed in Section~\ref{sec:simulations}. Section \ref{sec:conclusion} gives a concluding discussion. Our code, together with the traffic detection data and the Appendix is publicly available on GitHub\footnote{\url{https://github.com/luetkeschwienhorst/dropoutinextendedGLMs}}.

\section{Generalized Linear Models based on Double Exponential Families}
\label{sec:prelim}

\subsection{Double exponential family}
\label{subsec:def}

The distribution of a random vector $Y$ is in a natural exponential family (EF) if its target density has the form
\begin{equation}
\label{eq:ef}
f_\theta(y)=\exp\left(\langle\theta,y\rangle-b(\theta)+c(y)\right),
\end{equation}
where $c(y)=\log h(y)$ and $b(\theta)=\log C(\theta)^{-1}$ for known functions $C(\cdot)$ and $h(\cdot)$. The EF is parameterized through $\theta\in\Theta$. When $Y$ is a random variable with density \eqref{eq:ef}, $\mu:=\mathbb{E}_\theta[Y]=b'(\theta)$ and $\mathbb{V}_\theta[Y]=b''(\theta)$, thast is,
\begin{align*}
\mathbb{V}_{\theta}[Y]=\frac{\partial}{\partial \theta}\mathbb{E}_{\theta}[Y].
\end{align*}
This shows that the variance is determined by the mean function only. Many popular distributions such as the Gaussian, Poisson and binomial have densities of the form \eqref{eq:ef}. As discussed earlier, the implied mean-variance relationship for binomial and Poisson models may be inappropriate for real data, since for the Poisson distribution $\mathbb{V}_\theta(Y)=\mathbb{E}_\theta(Y)$, and for a binomial distribution with $N$ trials $\mathbb{V}_\theta(Y)=\mathbb{E}_\theta(Y)(N-\mathbb{E}_\theta(Y))/N$. For a general introduction and more details on EFs we refer to \citet{LehCas1998}.

\citet{Efr1986} extends the notion of the EF to the double exponential family (DEF) by introducing a dispersion parameter $\gamma>0$. Target densities in the DEF are of the form
\begin{align}
\label{eq:density_def}
f_{\gamma,\theta}(y):=C(\gamma,\theta)\gamma^{\frac{1}{2}}f_\theta(y)^\gamma f_{\theta(y)}(y)^{1-\gamma},
\end{align}
where $f_\theta$ and $f_{\theta(y)}$ are densities as in \eqref{eq:ef}, $C$ is a normalizing constant and $\theta(y)=(b')^{-1}(y)$ is the value of $\theta$ for which the mean is $y$, and we allow for ${\theta(y)\in \mathbb{R}\cup \{-\infty,\infty\}}$ to deal with boundary cases. The introduction of the dispersion parameter $\gamma$ decouples the mean and variance of the underlying natural EF and \citet{Efr1986} shows that for a random variable $Y\sim\text{DEF}(\theta,\gamma)$ with density \eqref{eq:density_def}, $\mathbb{E}[Y]\approx\mu$, $\mathbb{V}[Y]\approx b''(\theta)/\gamma$ and $C(\gamma,\theta)\approx 1$;  see \citet{Efr1986} for further discussion of the accuracy of these approximations. From the expression for the variance, $\gamma<1$ ($\gamma>1$)  corresponds to overdispersion (underdispersion). Also, we will assume that the map $\theta\mapsto b'(\theta)$ is one-to-one, such that we can parameterize via $\mu$ instead of $\theta$. We will write $f_{\gamma,\mu}$ and $Y\sim\text{DEF}(\mu,\gamma)$ instead of $f_{\gamma,\theta}$ and $Y\sim\text{DEF}(\theta,\gamma)$.

\subsection{GLMs based on DEFs}
\label{subsec:glms_based_on_defs}
Consider observed data $\{(y_i,\bm{x}_i,\bm{z}_i)\}_{i=1}^n$ of responses $y_i\in\mathcal{Y}\subseteq\mathbb{R}$ and feature vectors $\bm{x}_i\in\mathbb{R}^{d_\mu}$ and $\bm{z}_i\in\mathbb{R}^{d_\gamma}$. Conditionally on the features, the responses will be modelled as observations of independent random variables $Y_i$ with distributions from a DEF.  The features $\bm{x}_i$ and $\bm{z}_i$ will appear in models for the mean and dispersion respectively. In the regression context it is convenient to rewrite the natural EF target density \eqref{eq:ef} for scalar $Y_i$ as
\begin{align}
\label{regression_nef}
f_{\theta,\phi/\nu_i}(y_i) & =\exp\left(\frac{\theta_i y_i-b(\theta_i)}{\phi/\nu_i}+c\left(y_i,\phi/\nu_i\right)\right),
\end{align}
where $\phi$ is a fixed scale parameter, $\nu_i$ is a known weight and both $\nu_i$ and $\theta_i$ can vary between observations. The mean for density \eqref{regression_nef} is
$\mu_i=b'(\theta_i)$, and the variance is $(\phi/\nu_i)b''(\theta_i)$. For a conventional GLM with a binomial response such as a logistic regression, the weight $\nu_i$ would be $n_i$, the number of binomial trials for the $i$th observation. For binomial and Poisson GLMs the scale $\phi$ is $1$, but in a Gaussian linear regression
the scale parameter $\phi$ is the variance of the response.

We leave dependence of all quantities on $\phi/\nu_i$ implicit in our notation in the following discussion, retaining our previously established notation for DEFs. In our extended GLMs based on DEFs, we assume
\begin{align*}
Y_i\,|\,\theta_i,\gamma_i \sim\text{DEF}(\theta_i,\gamma_i),
\end{align*}
where, as discussed below, $\theta_i$ and $\gamma_i$ are functions of $\bm{x}_i$ and $\bm{z}_i$ respectively. Following \citet{Efr1986}, the DEF assumption implies
\begin{align*}
\mu_i&:=\mathbb{E}[Y_i\,|\,\theta_i,\gamma_i]\approx b'(\theta_i),\\
\sigma^2_i&:=\mathbb{V}[Y_i\,|\,\theta_i,\gamma_i]\approx \frac{\phi b''(\theta_i)}{\gamma_i \nu_i},
\end{align*}
and $C(\gamma_i,\theta_i)\approx 1$.  We take $\mu_i$ and $\gamma_i$ to be functions of linear predictors $\bm{x}_i^T \bm{\beta}$ and $\bm{z}_i^T \bm{\alpha}$, where $\bm{\beta}\in\mathbb{R}^{d_\mu}$ and $\bm{\alpha}\in\mathbb{R}^{d_\gamma}$ are unknown coefficient vectors. We choose a canonical link in the mean and a log-link for the dispersion so that
\begin{align*}
\theta_i=(b')^{-1}(\mu_i) &=\bm{x}_i^T\bm{\beta},\\
\log(\gamma_i)&=\bm{z}_i^T\bm{\alpha}.
\end{align*}
The log-link for the dispersion ensures the dispersion parameter is nonnegative. 

\section{Dropout regularization in GLMs based on DEFs}
\label{sec:dropout_mean_disp}

\subsection{Dropout regularization}
\label{subsec:dropout}
Dropout regularization randomly perturbs the observed feature vectors. Given some noise vector $\bm{\xi}$ and a noise function $\nu$, an observed feature vector $\bm{x}\in \mathbb{R}^d$ is transformed into $\widetilde{\bm{x}}:=\nu(\bm{x},\bm{\xi})$. 
The random perturbation is  unbiased, which means that $\mathbb{E}(\widetilde{\bm{x}})=x$. In what follows we use multiplicative noise, $\nu(\bm{x},\bm{\xi})=\bm{x}\odot\bm{\xi}$, where $\odot$ is the elementwise product of two vectors. Assuming $\mathbb{E}[\bm{\xi}]=\bm{1}_d$, we have $\mathbb{E}[\widetilde{\bm{x}}]=\bm{x}\odot \mathbb{E}[\bm{\xi}]=\bm{x}$. 

Typical choices for the distribution of the noise vector $\bm{\xi}$ include Bernoulli dropout with i.i.d. ${\xi_{j}\sim(1-\delta)^{-1}\text{Bernoulli}(1-\delta)}$, where $\delta\in (0,1)$ is the dropout probability, and Gaussian dropout with i.i.d. $\xi_{j}\sim\mathcal{N}(1,\sigma^2)$, with noise variance $\sigma^2$ for $j=1\ldots,d$. We have described the process of random perturbation for a single feature vector. With many feature vectors, different random perturbations are performed independently for each one. Extended GLMs incorporate features in both the mean and dispersion models and these need not be the same, although they can be.  When dropout is performed in both the mean and dispersion models, the perturbations will be independent in the mean and dispersion components.  When the features enter into a model linearly, perturbing the features is equivalent to perturbing the parameters independently in the terms of the loss function, since $(\bm{x}\odot \bm{\xi})^T \bm{\beta}=\bm{x}^T (\bm{\beta} \odot \bm{\xi})$.

\subsection{Dropout regularization for the mean parameter} \label{sec: Dropout regularization for the mean parameter}

We first consider dropout for extended GLMs, where dropout is performed only in the mean  and not the dispersion model. The argument closely follows the one given in 
\citet{WagWanLia2013} for the case of conventional GLMs. Dropout in both the mean and dispersion is more complex and is considered in the next subsection. 

Based on the assumptions from Subsection \ref{subsec:glms_based_on_defs}, dropout regularization in the mean model leads to the optimization problem
\begin{align}
\label{eq:mean_objective}
\min_{\bm{\beta},\bm{\alpha}}\sum_{i=1}^n-\mathbb{E}
\left[\ell_i\left(\bm{\beta}\odot\bm{\xi}_i,\bm{\alpha}\right)\right], &
\end{align}
where we have written $\ell_i(\beta,\alpha)$ for the log-likelihood term for the $i$th
observation, and the expectation is taken with respect to the distribution of the i.i.d.~vectors $\bm{\xi}_i$ with $\mathbb{E}[\bm{\xi}_i]=\bm{1}_{d_\mu}$ and $\mathbb{V}[\bm{\xi}_i]=\sigma^2_\mu\mathbb{I}_{d_\mu}$. Using the definition of the DEF, and assuming that $C(\gamma_i,\theta_i)=1$, 
\begin{align}
\label{eq:loglikelihood}
\ell_i(\bm{\beta},\bm{\alpha})&=\frac{1}{2}\bm{z}_i^T\bm{\alpha}+\exp(\bm{z}_i^T\bm{\alpha})
\frac{\left\{y_i\bm{x}_i^T\bm{\beta}-b(\bm{x}_i^T\bm{\beta})\right\}}{\phi/\nu_i}+(1-\exp(\bm{z}_i^T\bm{\alpha}))
\frac{\left\{y_i\theta(y_i)-b(\theta(y_i))\right\}}{\phi/\nu_i}.
\end{align}

Appendix \ref{app:dropout_mean} shows that \eqref{eq:mean_objective} is approximately equal to
\begin{align}
\label{eq:mean_objective_transformed2}
\min_{\bm{\beta},\bm{\alpha}}\left\{-\sum_{i=1}^n\ell_i(\bm{\beta},\bm{\alpha})+\frac{1}{2}\sigma^2_\mu\|\bm{\Theta}\bm{\beta}\|_2^2\right\},
\end{align}
where $\bm{\Theta}=\text{diag}(\bm{X}^T\bm{W}\bm{X})^{1/2}$ is a penalty matrix, $\bm{X}\in\mathbb{R}^{n,d_\mu}$ is the design matrix with $i$th row $\bm{x}_i$, and the diagonal weight matrix $\bm{W}$ depends on both $\bm{\beta}$ and $\bm{\alpha}$,
\begin{align}
\label{eq:weight_matrix}
\bm{W}:=\text{diag}\left(\frac{\gamma_1b''(\bm{x}_1^T\bm{\beta})}{\phi/\nu_1},\ldots,\frac{\gamma_nb''(\bm{x}_n^T\bm{\beta})}{\phi/\nu_n}\right)\in\mathbb{R}^{n,n}.
\end{align}
The L2 penalty shrinks the normalized vector $\bm{\Theta}\bm{\beta}$ towards the origin. As a result, the estimated weights $\hat{\beta}_j$ of some features can be close to zero, effectively removing the feature. The hyperparameter $\sigma^2_\mu$ controls the degree of shrinkage.

The form of $\bm{\Theta}$ allows us to understand which features will experience little penalty. The diagonal entries are
\begin{equation}
\label{eq:penalty_matrix_entry}
\Theta_{jj}=\left(\sum_{i=1}^n (\gamma_i\nu_i/\phi)^2\mathbb{V}[Y_i]x_{ij}^2\right)^{1/2}=\left(\sum_{i=1}^n(\nu_i/\phi)^2 \gamma_i \varphi_ix_{ij}^2\right)^{1/2},
\end{equation}
where $\mathbb{V}[Y_i]:= \phi b''(\bm{x}_i^T\bm{\beta})/(\nu_i\gamma_i)=\varphi_i/\gamma_i$ is the variance of the dependent variable $Y_i$ according to the model and $\varphi_i:=\phi b''(\bm{x}_i^T\bm{\beta})/\nu_i$ is the ``baseline'' variance when there is no over- or under-dispersion ($\gamma_i=1$). The expression inside the brackets in \eqref{eq:penalty_matrix_entry} is a rescaled second moment of the $j$th feature. It will be small if $x_{ij}$ is small -- or even zero -- for most $i$, or if $(\gamma_i\nu_i/\phi)^2\mathbb{V}[Y_i]=(\nu_i/\phi)^2\gamma_i\varphi_i$ is small enough when $x_{ij}$ is large. Thus, ``rare'' features which are close to zero for most samples, and which are associated with small baseline variances $\varphi_i$ and large
overdispersions $\gamma_i$ when the feature value is large, will experience little penalty. 

Our general perspective includes the special case of GLMs where $\gamma_i=1$ for $i=1,\dots, n$. Logistic regression was discussed in detail by \citet{WagWanLia2013} and they point out that little penalty is exerted on rare features which produce confident predictions. In addition, \citet{WagWanLia2013} state that $(1/n)\bm{X}^T\bm{W}\bm{X}$ is the observed Fisher information with respect to $\bm{\beta}$. This adds a geometric perspective to dropout regularization: the normalization of $\bm{\beta}$ by $\bm{\Theta}$ ensures that penalization is performed in accordance with the curvature of $\ell$ around the true parameter value. $\bm{\Theta}\bm{\beta}$ represents $\bm{\beta}$ in another basis  such that the level sets of $\ell$ parameterized in $\bm{\Theta}\bm{\beta}$ are spherical. We derive this Fisher information matrix in Appendix \ref{app:fisher_info}.

\subsection{Dropout regularization for the mean and dispersion parameter}
\label{subsec:dropout_mean_disp}

Next we extend to the case where dropout is performed in both the mean and dispersion models. As before, $\theta_i=\bm{x}_i^T\bm{\beta}$ and $\log(\gamma_i)=\bm{z}_i^T\bm{\alpha}$. Dropout regularization leads to the minimization problem
\begin{align}
\label{eq:mean_disp_objective}
\min_{\bm{\beta},\bm{\alpha}}
\sum_{i=1}^n-\mathbb{E}[\ell_i(\bm{\beta}\odot\bm{\xi}_i,\bm{\alpha}
\odot\bm{\zeta}_i)],
\end{align}
where the expectation is taken with respect to the dropout noise vectors $\bm{\xi}_i$ and $\bm{\zeta}_i$, which are independent
from each other. 
It is assumed that $\mathbb{E}[\xi_i]=\bm{1}_{d_\mu}$,
$\mathbb{V}[\xi_i]=\sigma_\mu^2 \mathbb{I}_{d_\mu}$, 
$\mathbb{E}[\zeta_i]=\bm{1}_{d_\gamma}$ 
and 
$\mathbb{V}[\zeta_i]=\sigma_\gamma^2 \mathbb{I}_{d_\gamma}$.  
Write $\widetilde{\theta}_i:=\bm{x}_i^T(\bm{\beta}\odot\bm{\xi}_i)$ and $\log(\widetilde{\gamma}_i):=\bm{z}_i^T(\bm{\alpha}\odot\bm{\zeta}_i)$.  Our assumptions on the dropout noise imply that
$\mathbb{E}[\widetilde{\theta}_i]=\theta_i$ and
$\mathbb{E}[\log \widetilde{\gamma}_i]=\log \gamma_i$.

In order to get a better understanding of dropout regularization in the dispersion, we aim to find an approximation of \eqref{eq:mean_disp_objective} similar to the one in \eqref{eq:mean_objective_transformed2}. We make a normality assumption, $\bm{\zeta}_i\sim\mathcal{N}(\bm{1}_{d_\gamma},\sigma^2_\gamma\mathbb{I}_{d_\gamma})$ so that
\begin{align} \label{eq: log_gamma_normal_approx}
\log(\widetilde{\gamma}_i)=\bm{z}_i^T(\bm{\alpha}\odot\bm{\zeta}_i)\sim\mathcal{N}\left(\bm{z}_i^T\bm{\alpha},\sigma^2_\gamma	\sum_{j=1}^{d_\gamma}z_{ij}^2\alpha_j^2\right).
\end{align}
Although this assumption may seem strong, \eqref{eq: log_gamma_normal_approx} will often 
be a good approximation for non-Gaussian dropout noise via a central limit argument.
Since $\widetilde{\gamma}_i$ is lognormal, its expectation is
\begin{align}
\mathbb{E}[\widetilde{\gamma}_i] &=\exp\left(\bm{z}_i^T\bm{\alpha}+\frac{1}{2}\sigma^2_\gamma	\sum_{j=1}^{d_\gamma}z_{ij}^2\alpha_j^2\right)
=\gamma_i\exp\left(\frac{1}{2}\sigma^2_\gamma\|\bm{z}_i\odot\bm{\alpha}\|_2^2\right). \label{expectation-log-gamma}
\end{align}
Using \eqref{eq:loglikelihood}, we obtain
\begin{align*}
\mathbb{E}[\ell_i(\bm{\beta}\odot \bm{\xi}_i, \bm{\alpha}\odot \bm{\zeta}_i]
&=
\frac{1}{2}\bm{z}_i^T\bm{\alpha}+\mathbb{E}[\widetilde{\gamma}_i]
\frac{\left\{y_i\bm{x}_i^T\bm{\beta}-\mathbb{E}[b(\bm{x}_i^T(\bm{\beta}\odot\bm{\xi}_i))]\right\}}{\phi/\nu_i}+(1-\mathbb{E}[\widetilde{\gamma}_i])
\frac{\left\{y_i\theta(y_i)-b(\theta(y_i))\right\}}{\phi/\nu_i}.
\end{align*}
Writing  $\bm{z}_i^T \bm{\alpha} =\log \mathbb{E}[\widetilde{\gamma}_i]-1/2\sigma_\gamma^2 \|\bm{z}_i\odot \bm{\alpha}\|^2,$ and
\begin{align*}
\widetilde{\ell}_i(\bm{\beta},\bm{\alpha})&= \frac{1}{2}\log \mathbb{E}[\widetilde{\gamma}_i]+\mathbb{E}[\widetilde{\gamma}_i]
\frac{\left\{y_i\bm{x}_i^T\bm{\beta}-b(\bm{x}_i^T\bm{\beta})\right\}}{\phi/\nu_i}+(1-\mathbb{E}[\widetilde{\gamma}_i])
\frac{\left\{y_i\theta(y_i)-b(\theta(y_i))\right\}}{\phi/\nu_i},
\end{align*}
and using a second-order Taylor series approximation of $\mathbb{E}[b(\bm{x}_i^T(\bm{\beta}\odot\bm{\xi}_i))]$ (see Appendix \ref{app:dropout_mean} for a derivation) gives the following approximation of \eqref{eq:mean_disp_objective}:
\begin{align}
\label{eq:mean_disp_objective_transformed}
\min_{\bm{\beta},\bm{\alpha}}\left\{\sum_{i=1}^n-
\widetilde{\ell}_i(\bm{\beta},\bm{\alpha})
+\frac{1}{2}\sigma^2_\mu\|\widetilde{\bm{\Theta}}\bm{\beta}\|_2^2+\frac{1}{4}\sigma^2_\gamma\|\bm{\Gamma}\bm{\alpha}\|_2^2\right\},
\end{align}
with the penalty matrices $\widetilde{\bm{\Theta}}=\text{diag}(\bm{X}^T\widetilde{\bm{W}}\bm{X})^{1/2}$ and $\bm{\Gamma}=\text{diag}(\bm{Z}^T\bm{Z})^{1/2}$ and the weight matrix
\begin{align}
\label{eq:noisy_weight_matrix}
\widetilde{\bm{W}}:=\text{diag}\left(\frac{\mathbb{E}[\widetilde{\gamma}_1]b''(\bm{x}_1^T\bm{\beta})}{\phi/\nu_1},\ldots,\frac{\mathbb{E}[\widetilde{\gamma}_n]b''(\bm{x}_n^T\bm{\beta})}{\phi/\nu_n}\right).
\end{align}
We will now address the behavior of all three terms in the approximation \eqref{eq:mean_disp_objective_transformed}.

\subsubsection{Misspecified log-likelihood}

The first term in \eqref{eq:mean_disp_objective_transformed} is the negative log-likelihood of a DEF model in which 
the $i$th observation has location parameter $\theta_i=\bm{x}_i^T\bm{\beta}$ and dispersion parameter
\begin{align*}
\mathbb{E}[\widetilde{\gamma}_i]=\gamma_i\exp\left(\frac{1}{2}\sigma^2_\gamma\|\bm{z}_i\odot\bm{\alpha}\|_2^2\right).
\end{align*}
If the originally specified model was correct, 
$\widetilde{\ell}_i(\bm{\beta},\bm{\alpha})$
is a misspecified log-likelihood term where
$\gamma_i$ is multiplied by a multiplicative factor $\exp(\frac{1}{2}\sigma^2_\gamma\|\bm{z}_i\odot\bm{\alpha}\|_2^2)$ 
which is larger than $1$. Hence, for the misspecified log-likelihood to achieve a similar fit to the correctly specified case, $\|\bm{z}_i\odot\bm{\alpha}\|_2^2$ has to be small favoring rare and important features. As $\bm{z}_i^T\bm{\alpha}\rightarrow-\infty$ implies $\gamma_i=\exp(\bm{z}_i^T\bm{\alpha})\rightarrow0$, this misspecification penalty favors overdispersion.   

\subsubsection{Penalty for the mean parameter}

The second term in \eqref{eq:mean_disp_objective_transformed} is a Tikhonov penalty on $\bm{\beta}$, where the penalty matrix $\widetilde{\bm{\Theta}}=\text{diag}(\bm{X}^T\widetilde{\bm{W}}\bm{X})^{1/2}$ is affected by the dropout noise in the dispersion model. Writing $\bm{\Gamma}_i:=\text{diag}(z_{i1},\ldots,z_{id_\gamma})$,
we define
\begin{align*}
\bm{\Lambda}:=\text{diag}\left(\exp(\sigma^2_\gamma\|\bm{\Gamma}_1\bm{\alpha}\|_2^2/2),\ldots,\exp(\sigma^2_\gamma\|\bm{\Gamma}_n\bm{\alpha}\|_2^2/2)\right)
\end{align*}
and then $\widetilde{\bm{W}}=\bm{\Lambda}\bm{W}$ due to $\mathbb{E}[\widetilde{\gamma}_i]=\gamma_i\exp((1/2)\sigma^2_\gamma\|\bm{\Gamma}_i\bm{\alpha}\|_2^2)$. The weight matrix in \eqref{eq:noisy_weight_matrix} is therefore a rescaled version of the weight matrix in \eqref{eq:weight_matrix}. Furthermore, $\sigma^2_\gamma>0$ and $\|\bm{\Gamma}_i\bm{\alpha}\|_2^2>0$ imply $\exp((1/2)\sigma^2_\gamma\|\bm{\Gamma}_i\bm{\alpha}\|_2^2)>1$, which yields the property $\widetilde{W}_{jj}>W_{jj}>0$. This carries over to the entries
\begin{align*}
\widetilde{\Theta}_{jj}=\left(\sum_{i=1}^n\exp\left((1/2)\sigma^2_\gamma\|\bm{\Gamma}_i\bm{\alpha}\|_2^2\right)(\nu_i/\phi)^2 \gamma_i\varphi_ix_{ij}^2\right)^{1/2}
\end{align*}
of $\widetilde{\bm{\Theta}}$, which then fulfill $\widetilde{\Theta}_{jj}>\Theta_{jj}>0$. The observations regarding $\Theta_{jj}$ from Section~\ref{sec: Dropout regularization for the mean parameter} apply to $\widetilde{\Theta}_{jj}$ as well, i.e.~rare but important features are favored and overdispersion can attenuate the penalty, such that locally the mean might be modelled by features which are not that rare. $\widetilde{\Theta}_{jj}$ contains the additional exponential term $\exp((1/2)\sigma^2_\gamma\|\bm{\Gamma}_i\bm{\alpha}\|_2^2)$ compared to $\Theta_{jj}$ increasing the penalty. This increase will be minimal however, if the terms $\|\bm{\Gamma}_i\bm{\alpha}\|_2^2=\|\bm{z}_i\odot\bm{\alpha}\|_2^2$ are negligible in size. 

\subsubsection{Penalty for the dispersion parameter}

The third term in \eqref{eq:mean_disp_objective_transformed} is also a Tikhonov penalty, but on $\bm{\alpha}$ and with the penalty matrix $\bm{\Gamma}=\text{diag}(\bm{Z}^T\bm{Z})^{1/2}$. It shrinks $\bm{\alpha}$ towards the origin, i.e. ~$\gamma_i=\exp(\bm{z}_i^T\bm{\alpha})\rightarrow1$ and therefore no overdispersion. For rare features the diagonal entries $\Gamma_{jj}=(\sum_{i=1}^nz_{ij}^2)^{1/2}$
are small, and then $\alpha_j$ can still be large.  Similar to the case for $\bm{\beta}$, the penalty for $\bm{\alpha}$ has an interpretation in terms of the Fisher information, with the observed Fisher information with respect to $\bm{\alpha}$ being approximately $(1/n)\bm{Z}^T\bm{Z}$; see Appendix \ref{app:fisher_info} for further details.\\

In summary, simultaneous dropout regularization on the mean and the dispersion parameter favors rare but important features both in the mean and the dispersion model. Over- or underdispersion can attenuate or strengthen the level of regularization in the mean. Further, the dropout noise in the dispersion model imposes an additional penalty in the mean, if deviations from the base variance cannot be modeled using rare features. While the degree of regularization in the dispersion is controlled by $\sigma^2_\gamma$ alone, it is regulated by $\sigma^2_\mu$ and $\sigma^2_\gamma$ together in the mean. Lastly, overdispersion appears to be favored relative to underdispersion for two reasons. First, if the features in the dispersion are not that rare, the large misspecification term will encourage overdispersion. Second, the increased penalization due to underdispersion could be large, such that modeling underdispersion might be avoided altogether.

Based on our analysis, an ``ideal scenario'' in which dropout regularization will be most successful is characterized by true parameter vectors $\bm{\beta}$ and $\bm{\alpha}$ which are sparse relating to a feature vector with only a few components having a significant impact. These significant features should be rare themselves, meaning they should be large only for a relatively small fraction of the data. In addition, we find that overdispersion, i.e.~the true dispersion function fulfills $\gamma(z)<1$ for all $z$, can be better handled by dropout compared to underdispersion. Further insights on the finite sample properties of employed approximations used can be found in Section \ref{app: approximations} of the Appendix.

\section{Application to adaptive smoothing with B-splines}
\label{sec:simulations}

\subsection{Simulations}

We consider the non-linear regression model 
\begin{align}
\label{eq:model_simulation}
    y_i \mid x_i \sim \text{DEF}\left(f(x_i),g(x_i)\right),\qquad i=1,\dots,n,
\end{align}
with $x_i\in[0,1]$. The functional effects $f(\cdot)$ and $g(\cdot)$ are modelled using a B-spline basis expansion after a transformation by appropriate
link functions. We use B-splines with a relatively large number of knots, so that each basis functions is supported on only a small compact subset of the domain $[0,1]$. If the true effects are mostly flat, but vary rapidly on small sub-intervals of $[0,1]$, then the B-spline basis functions are rare but important features.

Estimates are obtained using SGD, and the choice of optimal hyperparameters $\left(\sigma_\mu^*, \sigma_\gamma^*\right)$ for the dropout noise is performed
via random search cross-validation~(CV) \citep{BerBen2012}. A detailed description of the algorithm is given in Appendix~\ref{app: algorithm}. We consider Bernoulli dropout as well as Gaussian dropout and compare the performance to the PMLE approach discussed in \citet{GijProCla2010}.

\begin{figure}[b]
\includegraphics[width=\textwidth]{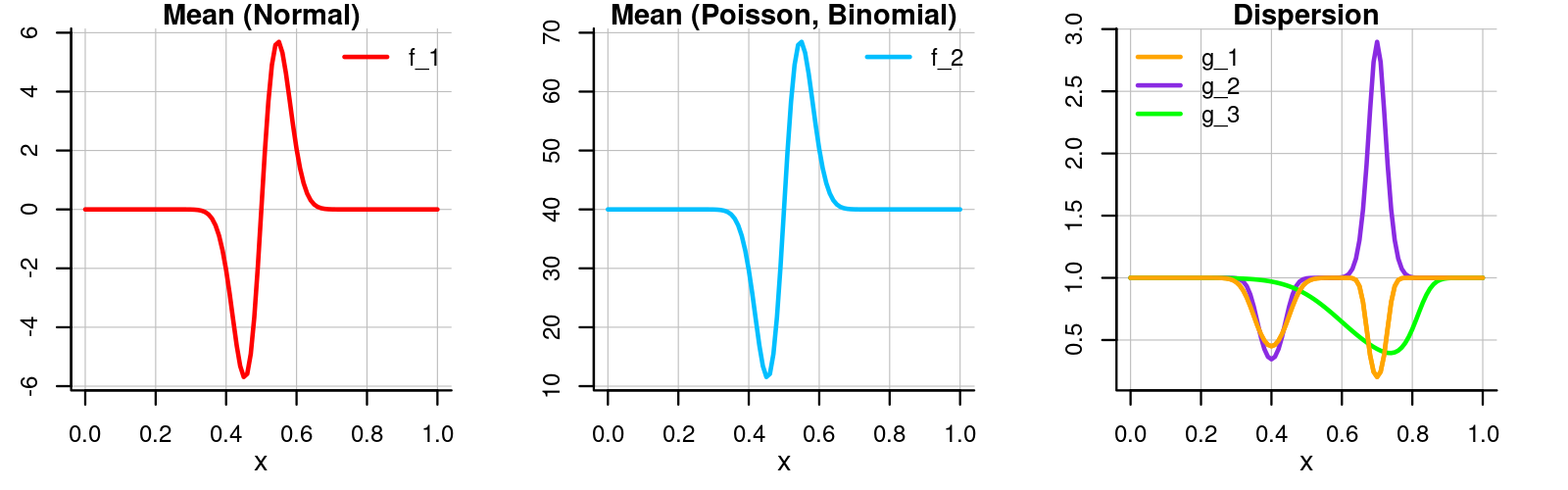}
\caption{Testfunctions for the mean and dispersion.}
\label{fig:testfunctions}
\end{figure}

We simulate data according to \eqref{eq:model_simulation} for the double Gaussian, double Poisson and double binomial distributions. For both dropout regularization and PMLE the mean is estimated well, but estimation of the dispersion function reveals differences in performance. Therefore, for each distribution we chose a similar mean function and paired it with three different dispersion functions in order to illustrate the conjectures from Section \ref{sec:dropout_mean_disp}.  A detailed description of the simulation design can be found in Appendix~\ref{app: simulation design}. The three dispersion testfunctions depicted in Figure \ref{fig:testfunctions} coincide with the subsequently mentioned Scenarios 1, 2 and 3. Additional results not presented in the main paper are given in Appendix~\ref{app:additional_plots}. The three different scenarios considered are:
\begin{itemize}
\item[] \textbf{Scenario 1:} The mean and dispersion functions are mostly flat, with rapid variation over some small intervals, and only overdispersion is present. This is a setting where dropout should perform well according to Subsection \ref{subsec:dropout_mean_disp}.
\item[] \textbf{Scenario 2:} Scenario 2 is similar to Scenario 1, but there is both over- and underdispersion.
\item[] \textbf{Scenario 3:} The mean function is mostly flat, with rapid variation over some small intervals, but the dispersion changes slowly with the feature values, and only overdispersion is present.
\end{itemize}
For each distribution and each Scenario we simulate $R=100$ replicate data sets with $n=250, 500, 1000$. To evaluate and compare different regularization methods we compute the root mean squared error (RMSE) over a fine equidistant grid on $[0,1]$.

\begin{figure}[t]
\centering
\begin{subfigure}{\textwidth}
\includegraphics[width=0.29\textwidth]{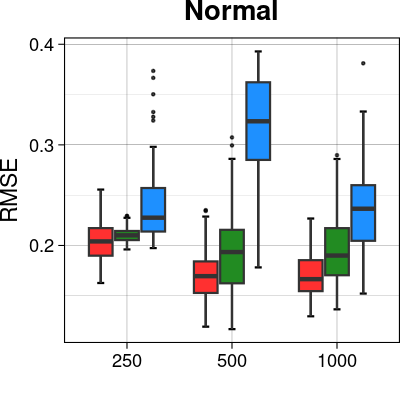}
\includegraphics[width=0.29\textwidth]{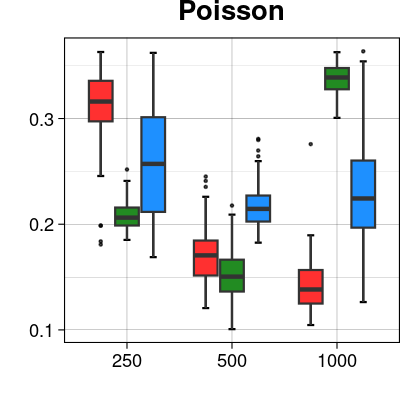}
\includegraphics[width=0.4\textwidth]{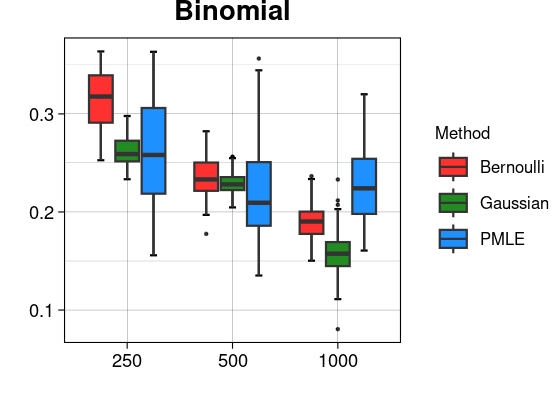}
\end{subfigure}
\begin{subfigure}{\textwidth}
\includegraphics[width=0.29\textwidth]{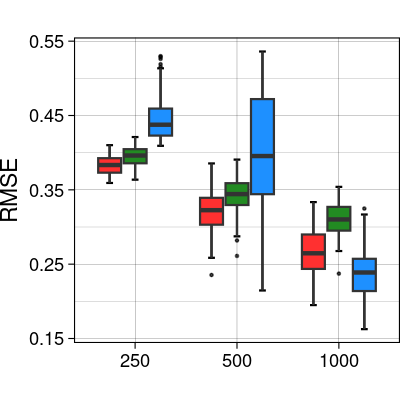}
\includegraphics[width=0.29\textwidth]{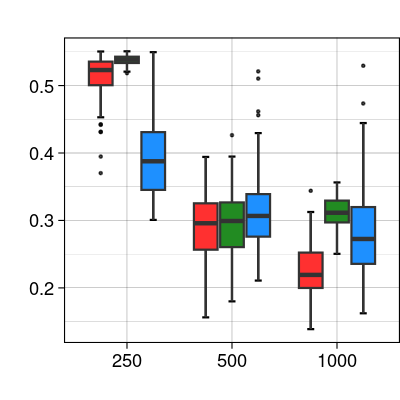}
\includegraphics[width=0.29\textwidth]{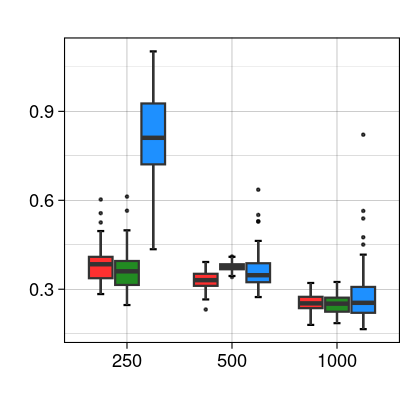}
\end{subfigure}
\begin{subfigure}{\textwidth}
\includegraphics[width=0.29\textwidth]{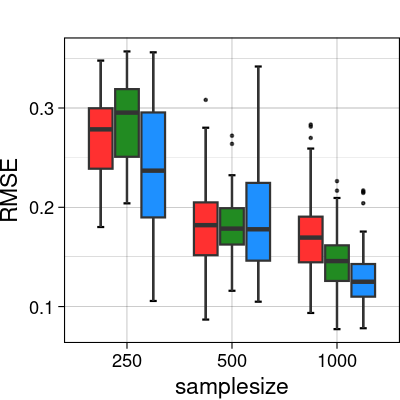}
\includegraphics[width=0.29\textwidth]{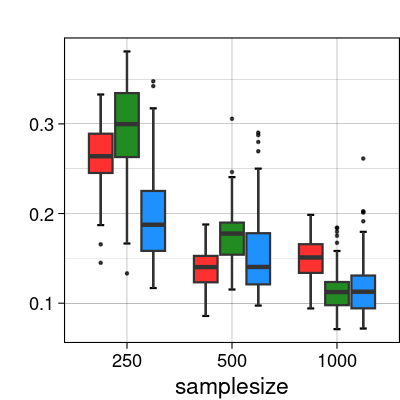}
\includegraphics[width=0.29\textwidth]{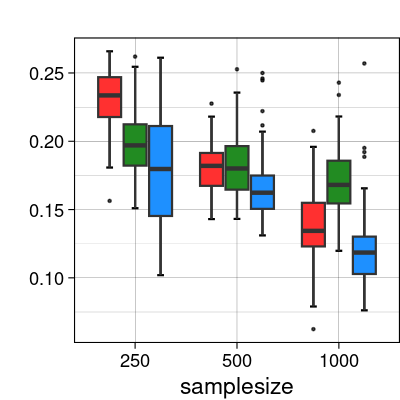}
\vspace{11pt}
\end{subfigure}
\caption{Boxplots of RMSEs for dispersion parameters of Gaussian data (left column), Poisson data (middle column) and binomial data (right column) across Scenarios 1 (top row), 2 (middle row) and 3 (bottom row). Outliers in the dispersion were cut at the $95$-th percentile.}
\label{fig:boxplots_disp}
\end{figure}

Figure~\ref{fig:boxplots_disp} presents boxplots for the RMSEs of the dispersion estimates for the three distributions and across all different scenarios. Figure~\ref{fig:boxplots_mean} in Appendix~\ref{app:additional_plots} contains the respective boxplots for the mean estimates. All three methods estimate the mean very well.  Figure~\ref{fig:boxplots_disp} reveals that dropout regularization is competitive with PMLE in all scenarios.

In Scenario 1 with Gaussian data, dropout outperforms PMLE across all sample sizes, and Bernoulli dropout performs slightly better than Gaussian dropout.  
With count data the difference between dropout and PMLE is 
less pronounced. 
Performance of Gaussian dropout is poor for Poisson data in Scenarios 1 and 2 for the largest sample size, $n=1000$.  
Figure~\ref{fig:effects_poisson_disp} gives evidence that this poor
performance is associated with large bias in estimation
of the dispersion function. 
However, Gaussian dropout performs slightly better than Bernoulli dropout in most cases involving count data.
In the binomial case, dropout shows improved performance
relative to PMLE as the sample size increases. 

Scenario 2 is based on rare but important features as well, but now for some values of the covariate there is strong underdispersion. 
We find that in small sample sizes dropout performs well compared
to dropout for Gaussian and binomial data, but poorly for Poisson
data.  In other cases the differences are minor.  
For Gaussian data, Bernoulli dropout is slightly better than Gaussian dropout, similar to Scenario 1.

Since the true dispersion function for Scenario 3 cannot be modeled through rare features, dropout performs slightly worse than PMLE across all distributions and sample sizes for this case. The findings with respect to the three scenarios match our theoretical analysis from Section~\ref{sec:dropout_mean_disp}.

\begin{figure*}[t]
\centering
\begin{subfigure}{0.49\textwidth}
\includegraphics[width=\textwidth]{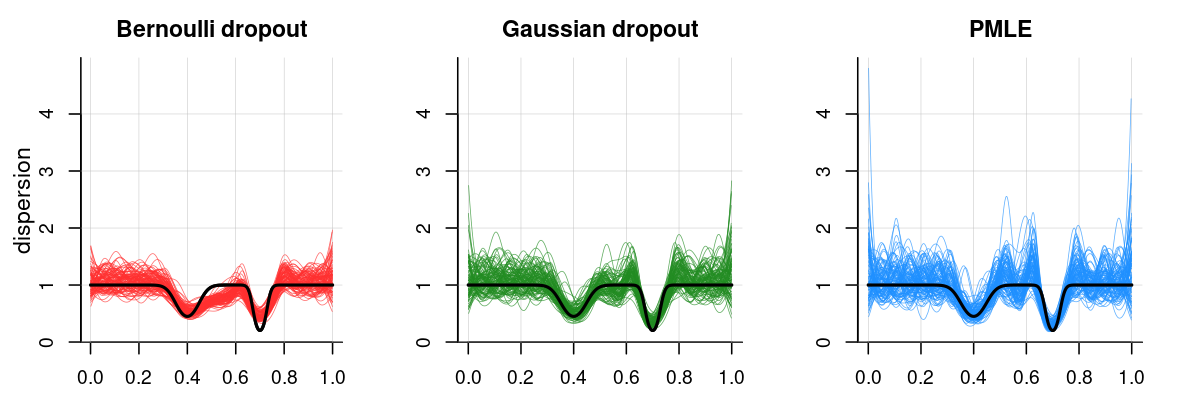}
\includegraphics[width=\textwidth]{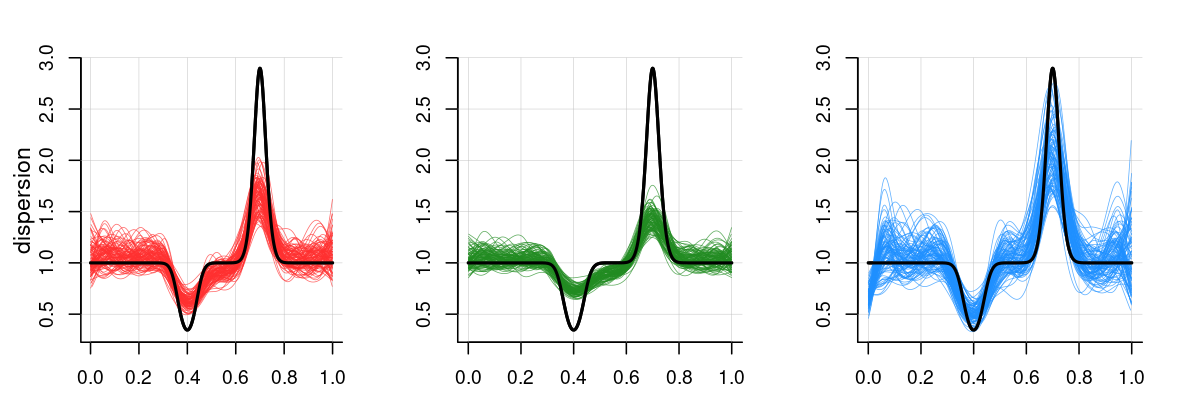}
\includegraphics[width=\textwidth]{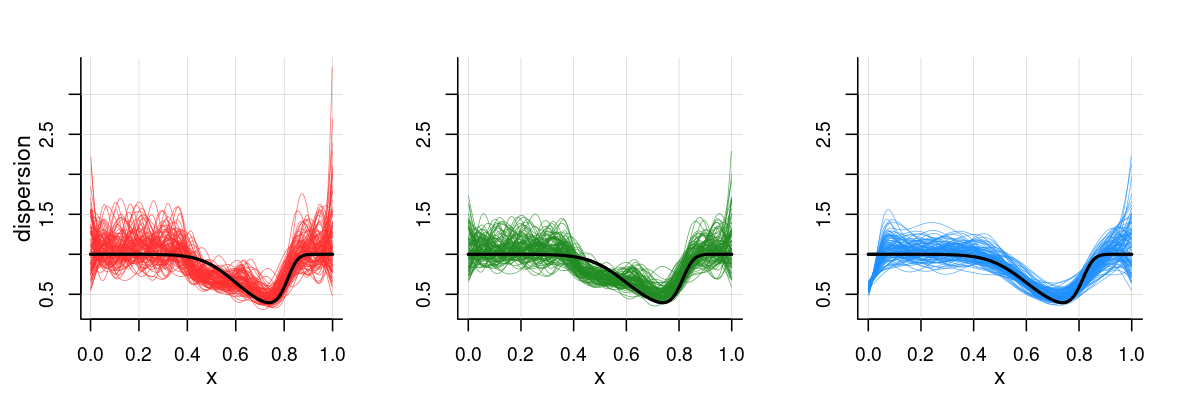}
\caption{Gaussian model}
\label{fig:effects_normal_disp}
\end{subfigure}
\begin{subfigure}{0.49\textwidth}
\includegraphics[width=\textwidth]{disp_comparison_normal_14_1000}
\includegraphics[width=\textwidth]{disp_comparison_normal_12_1000}
\includegraphics[width=\textwidth]{disp_comparison_normal_13_1000}
\caption{Binomial model}
\label{fig:effects_binomial_disp}
\end{subfigure}
\caption{Estimated dispersion effects in the (a) Gaussian model and  (b) binomial model for Bernoulli dropout (left), Gaussian dropout (middle) and PMLE (right) in Scenario 1 (upper row), Scenario 2 (middle row) and Scenario 3 (bottom row) for  $R=100$ replicates and $n=1,000$. The true effects are given by the black lines.}
\end{figure*}

Figure~\ref{fig:effects_normal_disp} depicts the estimated dispersion effects from Gaussian data for all three scenarios and $n=1,000$. Figure~\ref{fig:effects_normal_mean} in Appendix~\ref{app:additional_plots} contains the same plots for the mean estimates.  The estimation of the mean function is quite accurate across the three methods. The estimated dispersion functions are less accurate since estimation is more difficult \citep{GijProCla2010}. 
However, in most cases the dispersion estimates reproduced the correct shape. The PMLE estimates suffer from oscillations at the boundaries.
For Scenario 2, the rapid changes are not captured very well by Gaussian dropout, particularly where they correspond to regions
of underdispersion.  The problem also occurs with Bernoulli dropout but is less pronounced. For Scenario 3 the dropout estimates of the dispersion function are noticeably less smooth than the estimates obtained via PMLE. This is consistent with the boxplots at the bottom of the first column of Figure~\ref{fig:boxplots_disp}.

Figure~\ref{fig:effects_binomial_disp} presents the estimated dispersion effects obtained from binomial data. Previous observations with respect to Gaussian data apply as well. Additionally, we can observe a notable local minimum in 
the dispersion estimates in the vicinity of $x\approx0.5$. This is even more pronounced in case of smaller sample sizes (not shown).  It confirms the observation from Section~\ref{sec:dropout_mean_disp}, that there is an incentive to compensate a large variance caused by the mean with overdispersion. The variance function of the binomial distribution is given by $V(\mu)=\mu(N-\mu)/N$, i.e.~its maximizer is $N/2$. Thus, one can expect this behavior to take place at $x\approx 0.5$ for the mean function depicted in Figure A.1.  For the Poisson distribution, the variance function is given by $\mathbb{V}(\mu)=\mu$, and again referring to Figure A.1, we expect the regularization to favour overdispersion around $x\approx0.55$. This could be the reason for the reduced competitiveness of dropout regularization in the case of count data.   
The PMLE estimates are less smooth than the dropout estimates in Scenarios 1 and 2 and they oversmooth the true function in Scenario 3.

\subsection{Traffic detection data}
\label{sec:traffic_data}

\begin{figure}[!ht]
\centering
\begin{subfigure}{0.98\textwidth}
\includegraphics[width=\textwidth]{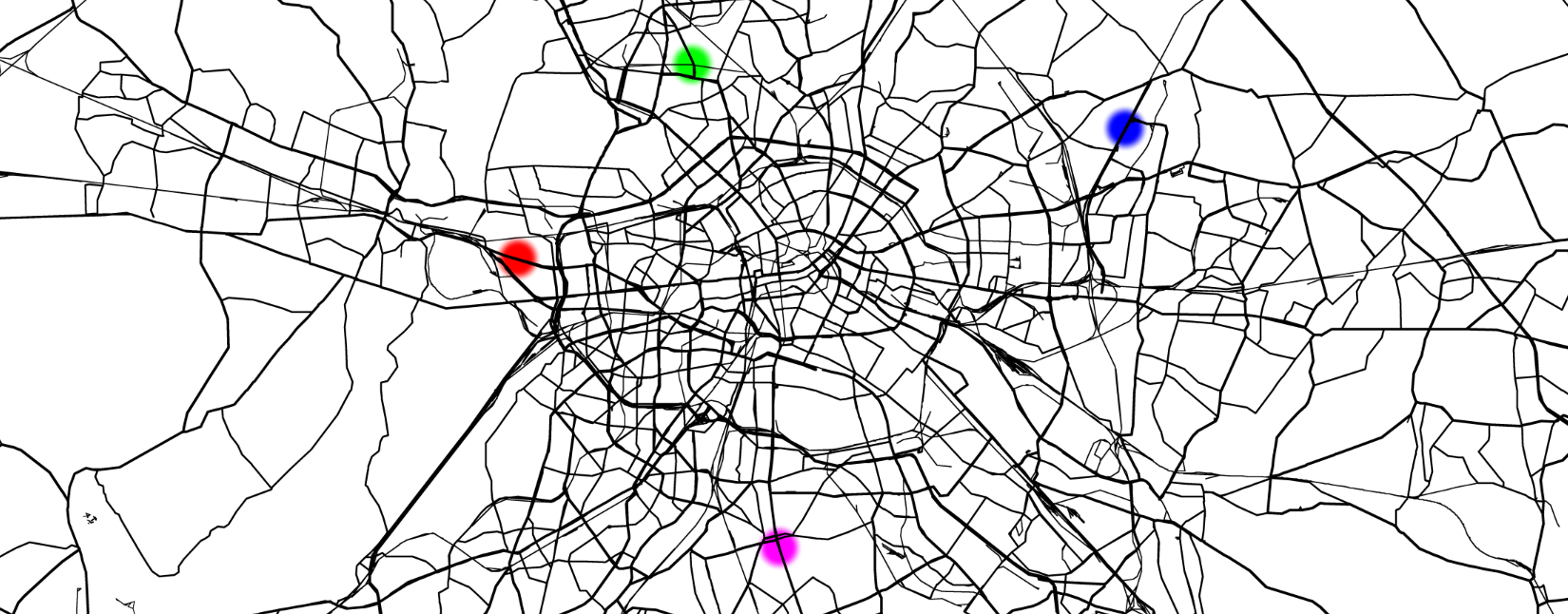}
\vspace{0mm}
\caption{Positioning of sensors}
\vspace{2mm}
\label{fig:traffic_data_map}
\end{subfigure}
\begin{subfigure}{0.24\textwidth}
\includegraphics[width=\textwidth]{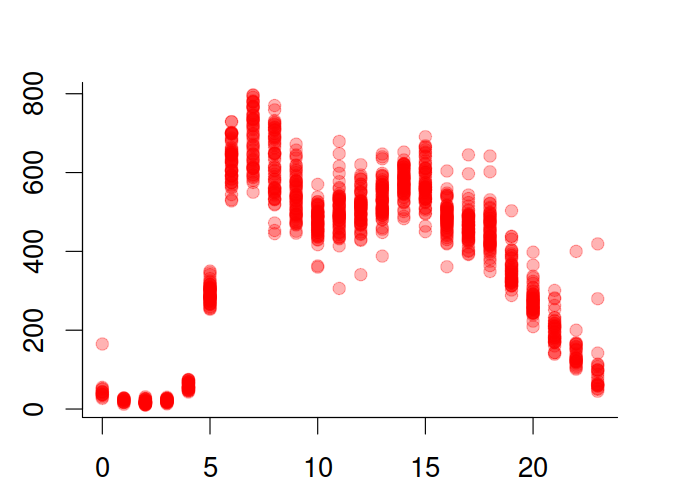}
\includegraphics[width=\textwidth]{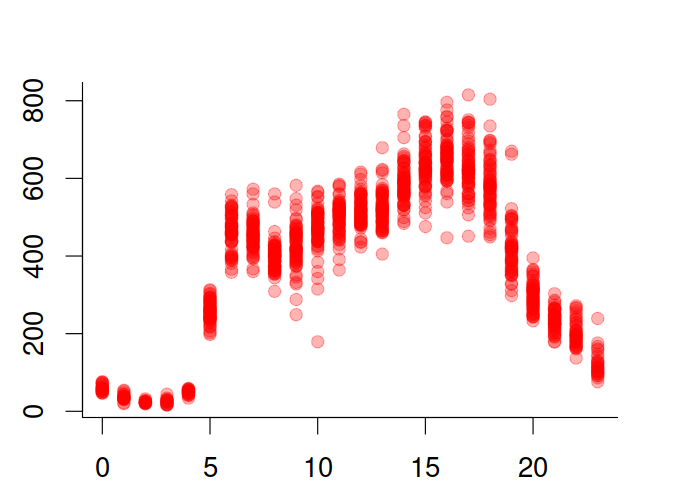}
\vspace{0mm}
\caption{West}
\label{fig:traffic_data_west}
\vspace{2mm}
\end{subfigure}
\begin{subfigure}{0.24\textwidth}
\includegraphics[width=\textwidth]{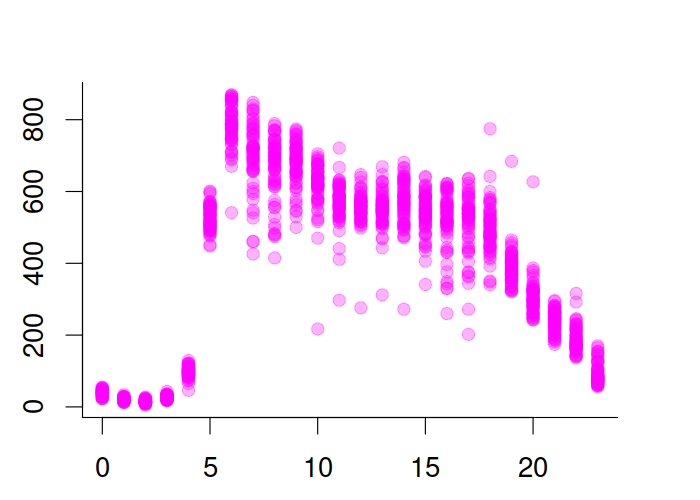}
\includegraphics[width=\textwidth]{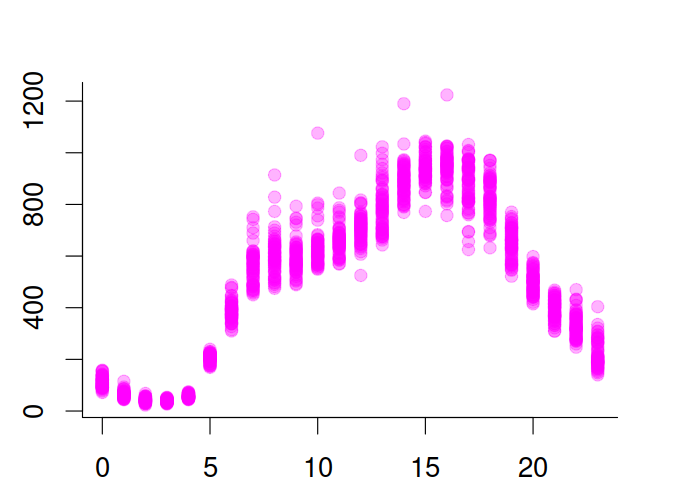}
\vspace{0mm}
\caption{South}
\vspace{2mm}
\end{subfigure}
\begin{subfigure}{0.24\textwidth}
\includegraphics[width=\textwidth]{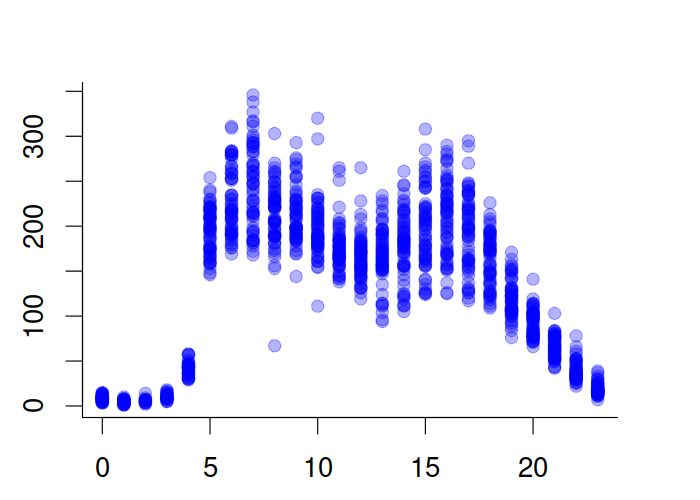}
\includegraphics[width=\textwidth]{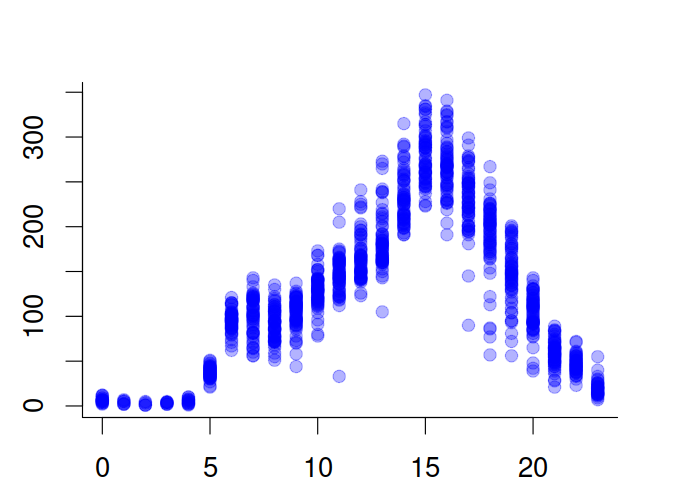}
\vspace{0mm}
\caption{East}
\vspace{2mm}
\end{subfigure}
\begin{subfigure}{0.24\textwidth}
\includegraphics[width=\textwidth]{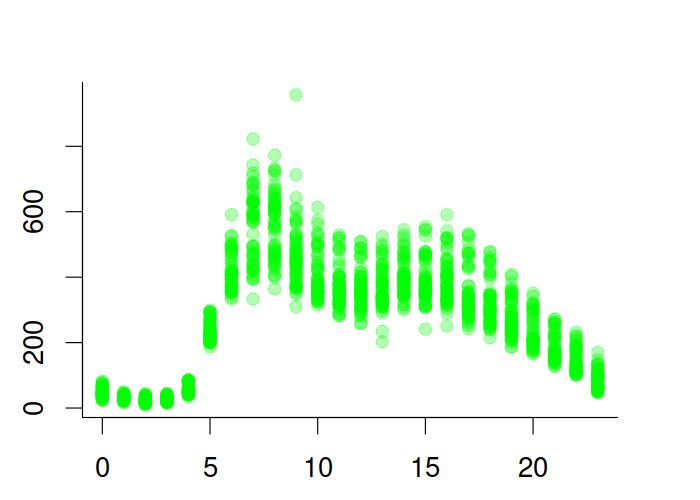}
\includegraphics[width=\textwidth]{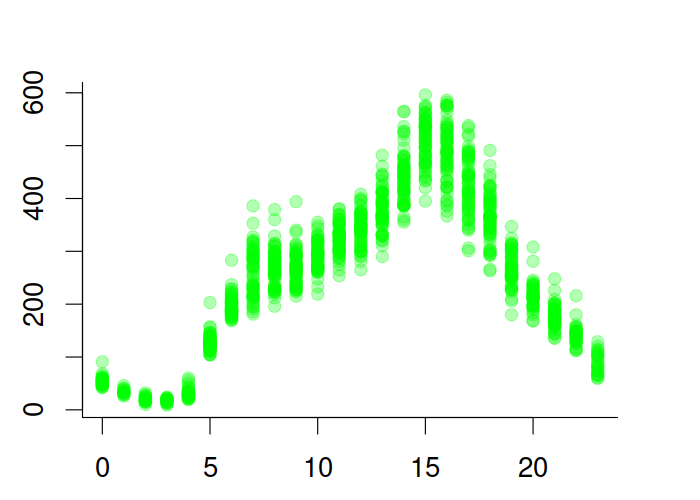}
\vspace{0mm}
\caption{North}
\vspace{2mm}
\label{fig:traffic_data_north}
\end{subfigure}
\caption{Traffic detection data with the (a) positioning of the four sensors in the (b) West, (c) South, (d) East and (e) North of the Berlin city center. For (b)--(e) the upper panels depict the counts ($y$-axis) for inbound traffic for each hour from 0am (=0) to 11pm (=23) ($x$-axis).  The bottom panels show the corresponding outbound traffic.}
\label{fig:traffic_data}
\end{figure}

Several hundred sensors monitor the traffic along main roads in the German capital Berlin. These sensors provide hourly aggregated data on the number of cars passing the sensors that is publicly available through the Senatsverwaltung für Umwelt, Mobilität, Verbraucher- und Klimaschutz and Verkehrsinformationszentrale Berlin\footnote{\url{https://viz.berlin.de}}. The Berlin traffic detection data was recently analysed in a regression context in \citet{KocKle2023}. Here, we consider data collected during summer of 2019 (June, July, and August)  from four distinct locations on the outskirts of the city center (see Figure~\ref{fig:traffic_data_map}), leading to a total of $10,314$ data points. The sensors are located at streets connecting outskirt neighbourhoods with the city center. Hence, as one might expect, the number of cars peaks during rush hour (see Figures~\ref{fig:traffic_data_west}--\ref{fig:traffic_data_north}), when people are commuting between home and work making this data set especially suitable for an analysis with a dropout DEF-GLM. 
\begin{figure}[h]
\centering
\begin{subfigure}{0.49\textwidth}
\includegraphics[width=\textwidth]{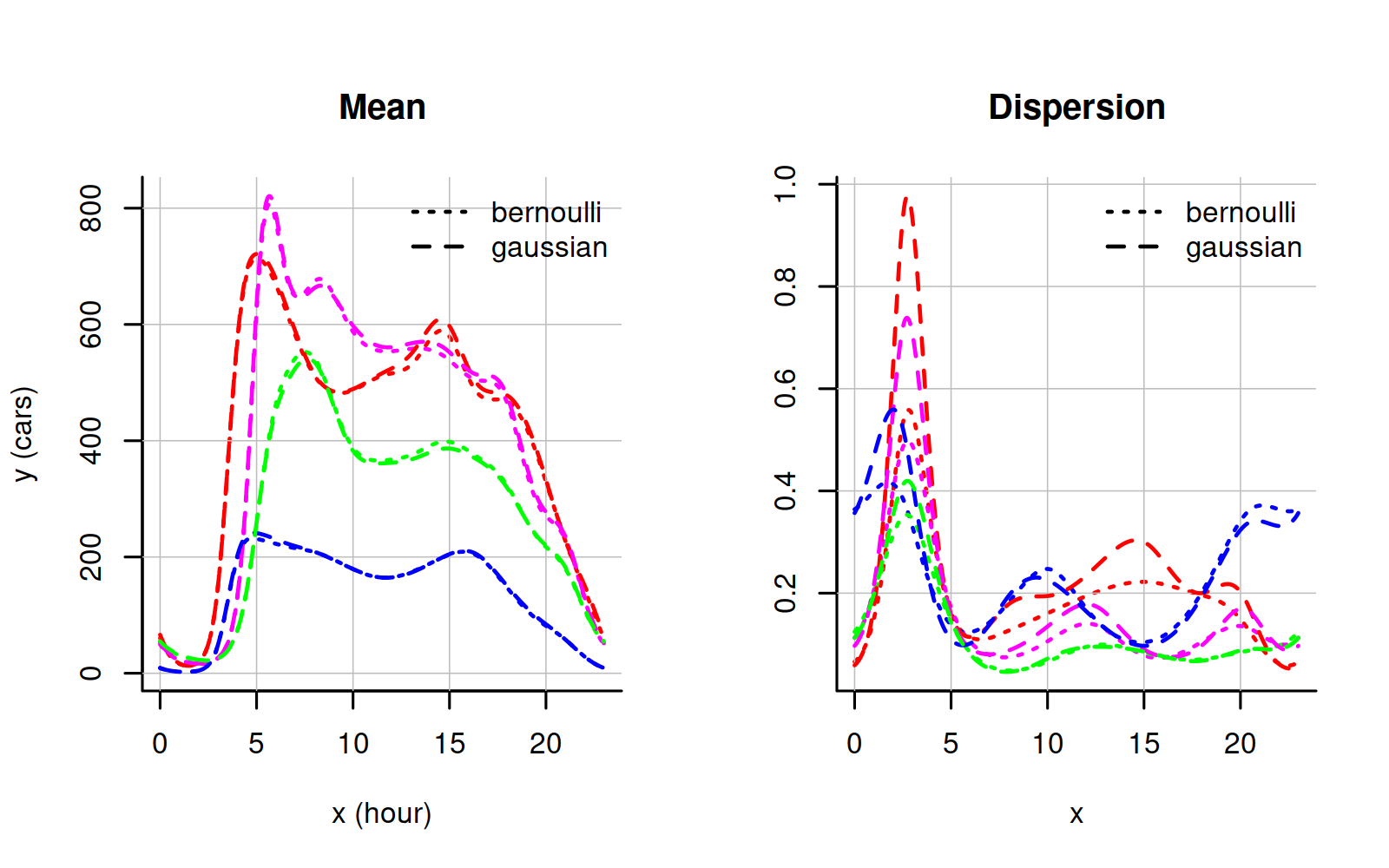}
\caption{Inbound}
\end{subfigure}
\begin{subfigure}{0.49\textwidth}
\includegraphics[width=\textwidth]{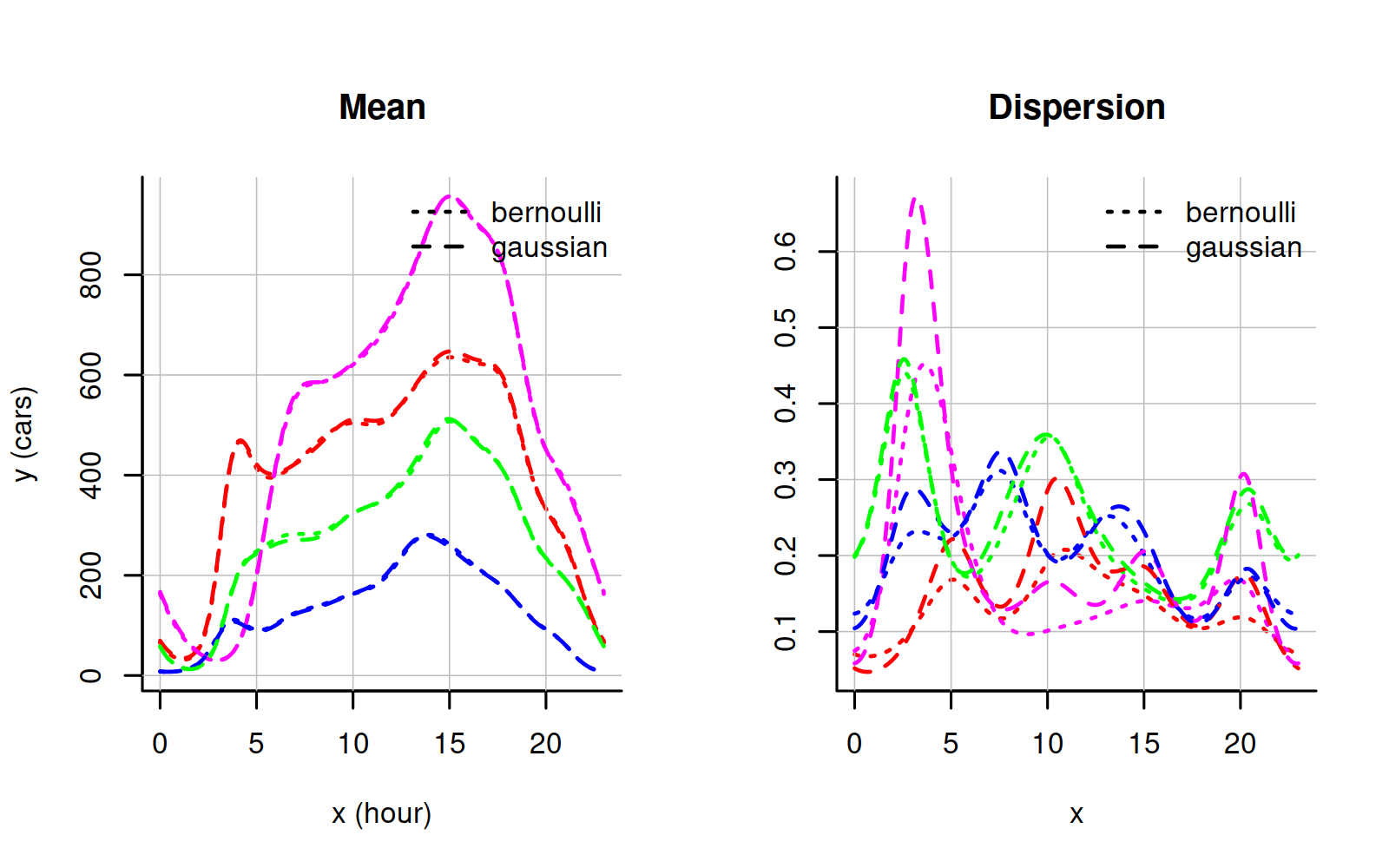}
\caption{Outbound}
\end{subfigure}
\caption{Cross-validated estimates for the traffic detection data of the four sensors in the ($\color{red}\bigcdot$) West, ($\color{magenta}\bigcdot$) South, ($\color{blue}\bigcdot$) East and ($\color{green}\bigcdot$) North of the Berlin city center.}
\label{fig:cv_estimates_traffic_data}
\end{figure}
Denote the number of cars passing the sensor at time-point $t$ as $y_t$. We assume a double Poisson model $y_t\mid t \sim \text{DPoi}\big(f(t),g(t)\big),$
where the functional effects $f(t)$ and $g(t)$ are effects of the day time $t$ in the mean and disperson modelled using cyclic B-spline basis expansions. The dropout rates ($p_\mu$, $p_\gamma$ and $\sigma_\mu$, $\sigma_\gamma$ for Bernoulli and Gaussian dropout, respectively) are chosen  via random search cross-validation by sampling 5,000 times from the sets $\mathcal{H}_{\text{Ber}}=[0,1]^2$ and $\mathcal{H}_{\mathcal{N}}=[0,2]^2$ (see Figure~\ref{fig:cv_traffic_data} in the Appendix).

Estimates based on the optimal dropout rates are shown in Figure~\ref{fig:cv_estimates_traffic_data}.  We find a similar behaviour across all four sensors. The mean effects are bimodal with one peak in the morning and a second peak in the evening. For the inbound traffic the peaks in the morning are more pronounced than the peaks in the afternoon and the outbound traffic mirrors this behaviour having a stronger peak in the evening. This corresponds well with an expected rush hour effect as people drive into the city for work in the morning and come back in the evening. These findings indicate that the dropout estimators are well suited to capture the peaks during morning and evening rush hours for inbound and outbound traffic respectively. Moreover, time-dependent overdispersion $(\gamma < 1)$ with narrow peaks in the early morning justifies the use of the extended GLM. The mean estimates obtained from Bernoulli noise are virtually indistinguishable from the ones based on Gaussian noise, which is most likely due to the small degree of regularization in the mean across all data sets (see Figure~\ref{fig:cv_traffic_data} in the Appendix). Differences are more pronounced between the dispersion estimates. However this is to be expected, since the dispersion is more difficult to estimate.

\section{Conclusion}
\label{sec:conclusion}

We studied dropout regularization in the context of flexible GLMs based on the DEF. Our theoretical analysis shows that
dropout favors rare but important features in the mean and  dispersion parameters, and overdispersion relative to underdispersion. Overdispersion alleviates the penalization on the mean provided the dispersion can be modelled by rare but important features itself. These findings are further justified by an empirical application to adaptive smoothing with B-splines and an application to real data on traffic detection in Berlin. Our experiments confirm that dropout regularization outperforms PMLE under ideal conditions. Deviations from the ideal scenario, such as the presence of underdispersion or a mean or dispersion function which cannot be modelled by rare but important features, leads to a decrease in performance of dropout regularization relative to PMLE. Thus, our findings extend previous work on dropout regularization for GLMs and add to the theoretical understanding of dropout methods in general.

However, the presented results can surely be broadened in a number of ways. Among them could be an extension of our findings to generalized additive models and quasi-likelihood estimation. Considering a greater number of real-world datasets certainly presents a challenge, given that the requirements on such data are quite particular, but it would unquestionably generalize our work. Lastly, the interpretable approximations used in our work to gain understanding of dropout regularization in extended GLMs are useful for predicting what we observe in the numerical experiments. Yet it must be acknowledged that we use approximations, and that there may be some situations where these approximations are inadequate. It could be interesting to investigate this more systematically in future work.

\begin{credits}
\subsubsection{\ackname} Nadja Klein acknowledges support through the Emmy Noether grant KL 3037/1-1 of the German research foundation (DFG). 
\end{credits}

%
%
%

\begin{thebibliography}{26}
\providecommand{\natexlab}[1]{#1}
\providecommand{\url}[1]{\texttt{#1}}
\providecommand{\urlprefix}{URL }
\expandafter\ifx\csname urlstyle\endcsname\relax
  \providecommand{\doi}[1]{doi:\discretionary{}{}{}#1}\else
  \providecommand{\doi}{doi:\discretionary{}{}{}\begingroup
  \urlstyle{rm}\Url}\fi

\bibitem[{Baldi and Sadowski(2013)}]{BalSad2013}
Baldi, P., Sadowski, P.J.: Understanding dropout. In: Advances in Neural
  Information Processing Systems, vol.~26 (2013)

\bibitem[{Bergstra and Bengio(2012)}]{BerBen2012}
Bergstra, J., Bengio, Y.: Random search for hyper-parameter optimization.
  Journal of Machine Learning Research \textbf{13}, 281--305 (2012)

\bibitem[{Bishop(1995)}]{Bis1995}
Bishop, C.M.: Training with noise is equivalent to tikhonov regularization.
  Neural Computation \textbf{7}(1), 108--116 (1995)

\bibitem[{Blei et~al.(2003)Blei, Ng, and Jordan}]{BleNgJor2003}
Blei, D.M., Ng, A.Y., Jordan, M.I.: Latent dirichlet allocation. Journal of
  machine Learning research \textbf{3}(Jan), 993--1022 (2003)

\bibitem[{Burges and Sch\"{o}lkopf(1996)}]{BurSch1996}
Burges, C.J.C., Sch\"{o}lkopf, B.: Improving the accuracy and speed of support
  vector machines. In: Advances in Neural Information Processing Systems,
  vol.~9 (1996)

\bibitem[{Dunn and Smyth(2018)}]{DunSmy2018}
Dunn, P.K., Smyth, G.K.: Generalized Linear Models With Examples in R. Springer
  Texts in Statistics, Springer New York, 1st edn. (2018)

\bibitem[{Efron(1986)}]{Efr1986}
Efron, B.: Double exponential families and their use in generalized linear
  regression. Journal of the American Statistical Association \textbf{81}(395),
  709--721 (1986)

\bibitem[{Eilers and Marx(1996)}]{EilMar1996}
Eilers, P.H.C., Marx, B.D.: Flexible smoothing with b-splines and penalties.
  Statistical Science \textbf{11}(2), 89--121 (1996)

\bibitem[{Gijbels et~al.(2010)Gijbels, Prosdocimi, and
  Claeskens}]{GijProCla2010}
Gijbels, I., Prosdocimi, I., Claeskens, G.: Nonparametric estimation of mean
  and dispersion functions in extended generalized linear models. {TEST}
  \textbf{19}(3), 580--608 (2010)

\bibitem[{Hannah et~al.(2011)Hannah, Blei, and Powell}]{HanBlePow2011}
Hannah, L.A., Blei, D.M., Powell, W.B.: Dirichlet process mixtures of
  generalized linear models. Journal of Machine Learning Research
  \textbf{12}(6) (2011)

\bibitem[{Helmbold and Long(2015)}]{HelLon2015}
Helmbold, D.P., Long, P.M.: On the inductive bias of dropout. Journal of
  Machine Learning Research \textbf{16}(105), 3403--3454 (2015)

\bibitem[{Hinton et~al.(2012)Hinton, Srivastava, Krizhevsky, Sutskever, and
  Salakhutdinov}]{HinSriNit2012}
Hinton, G.E., Srivastava, N., Krizhevsky, A., Sutskever, I., Salakhutdinov,
  R.R.: Improving neural networks by preventing co-adaptation of feature
  detectors. {arXiv}: 1207.0580 (2012)

\bibitem[{Kock and Klein(2023)}]{KocKle2023}
Kock, L., Klein, N.: Truly multivariate structured additive distributional
  regression. arXiv:2306.02711 (2023)

\bibitem[{Lee and Nelder(2000)}]{LeeNel2000}
Lee, Y., Nelder, J.A.: The relationship between double-exponential families and
  extended quasi-likelihood families, with application to modelling geissler's
  human sex ratio data. Journal of the Royal Statistical Society. Series C
  \textbf{49}(3), 413--419 (2000)

\bibitem[{Lehmann and Casella(1998)}]{LehCas1998}
Lehmann, E.L., Casella, G.: Theory of Point Estimation. Springer texts in
  statistics, Springer, 2nd edn. (1998)

\bibitem[{Maaten et~al.(2013)Maaten, Chen, Tyree, and
  Weinberger}]{MaaCheTzrWei2013}
Maaten, L., Chen, M., Tyree, S., Weinberger, K.: Learning with marginalized
  corrupted features. In: Proceedings of the 30th International Conference on
  Machine Learning, vol.~28, pp. 410--418, Atlanta, Georgia, USA (2013)

\bibitem[{{McCullagh} and Nelder(1998)}]{McCNel1998}
{McCullagh}, P., Nelder, J.A.: Generalized Linear Models. No.~37 in Monographs
  on Statistics and Applied Probability, Chapman \& Hall/{CRC}, 2nd edn. (1998)

\bibitem[{Merity et~al.(2017)Merity, Keskar, and Socher}]{MerKesNitSoc2017}
Merity, S., Keskar, N.S., Socher, R.: Regularizing and optimizing lstm language
  models. arXiv:1708.02182 (2017)

\bibitem[{Shao(2003)}]{Sha2003}
Shao, J.: Mathematical Statistics. Springer texts in statistics, Springer, 2nd
  edn. (2003)

\bibitem[{Srivastava et~al.(2014)Srivastava, Hinton, Krizhevsky, Sutskever, and
  Salakhutdinov}]{Srivastava+hkss14}
Srivastava, N., Hinton, G., Krizhevsky, A., Sutskever, I., Salakhutdinov, R.:
  Dropout: A simple way to prevent neural networks from overfitting. Journal of
  Machine Learning Research \textbf{15}(56), 1929--1958 (2014)

\bibitem[{Tan and Le(2019)}]{TanLe2019}
Tan, M., Le, Q.: {E}fficient{N}et: Rethinking model scaling for convolutional
  neural networks. In: Proceedings of the 36th International Conference on
  Machine Learning, vol.~97, pp. 6105--6114 (2019)

\bibitem[{Van~Erven et~al.(2014)Van~Erven, Kot{\l}owski, and
  Warmuth}]{VanKotWojWar2014}
Van~Erven, T., Kot{\l}owski, W., Warmuth, M.K.: Follow the leader with dropout
  perturbations. In: Proceedings of The 27th Conference on Learning Theory,
  vol.~35, pp. 949--974, Barcelona, Spain (2014)

\bibitem[{Wager et~al.(2013)Wager, Wang, and Liang}]{WagWanLia2013}
Wager, S., Wang, S., Liang, P.S.: Dropout training as adaptive regularization.
  In: Advances in Neural Information Processing Systems, vol.~26 (2013)

\bibitem[{Wang and Manning(2013)}]{WanMan2013}
Wang, S., Manning, C.: Fast dropout training. In: Proceedings of the 30th
  International Conference on Machine Learning, vol.~28, pp. 118--126, Atlanta,
  Georgia, USA (2013)

\bibitem[{Wei et~al.(2020)Wei, Kakade, and Ma}]{WeiKakTen2020}
Wei, C., Kakade, S., Ma, T.: The implicit and explicit regularization effects
  of dropout. In: Proceedings of the 37th International Conference on Machine
  Learning, vol. 119, pp. 10181--10192 (2020)

\bibitem[{Zeiler(2012)}]{Zei2012}
Zeiler, M.D.: {ADADELTA}: An adaptive learning rate method. {arXiv}:1212.5701
  (2012)

\end{thebibliography}
%


\newpage
\onecolumn

\appendix
\section{Appendix}

\subsection{Transformation of the dropout objective}
\label{app:dropout_mean}

Recall that $\mathbb{E}[\bm{x}_i^T(\bm{\beta}\odot\bm{\xi}_i)]=\bm{x}_i^T\bm{\beta}$. Writing $\mathbb{E}[b(\bm{x}_i^T(\bm{\beta}\odot\bm{\xi}_i))]=\mathbb{E}[b(\bm{x}_i^T(\bm{\beta}\odot\bm{\xi}_i))]+b(\bm{x}_i^T\bm{\beta})-b(\bm{x}_i^T\bm{\beta})$ the objective in 
\eqref{eq:mean_objective} is
\begin{align}
\label{eq:mean_objective_transformed}
\min_{\bm{\beta},\bm{\alpha}}\left\{-\sum_{i=1}^n\ell_i(\bm{\beta},\bm{\alpha})+\sum_{i=1}^n\gamma_i
\frac{\left\{\mathbb{E}[b(\bm{x}_i^T(\bm{\beta}\odot\bm{\xi}_i))]-b(\bm{x}_i^T\bm{\beta})\right\}}{\phi/\nu_i}\right\}
\end{align}
This shows that the original dropout objective is the maximum likelihood objective for $\bm{\beta}$ and $\bm{\alpha}$ plus a penalization term. The penalization term in \eqref{eq:mean_objective_transformed} does not depend on the responses $y_i$ and therefore does not penalize the accuracy. The convexity of $b$ together with Jensen's inequality imply that
\begin{align}
\label{eq:jensen}
\mathbb{E}[b(\bm{x}_i^T(\bm{\beta}\odot\bm{\xi}_i))]-b(\bm{x}_i^T\bm{\beta})\geqslant 0,
\end{align}
i.e.~nonnegativity of the penalty.

Given some random variable $X$ and a measurable function $f$, 
we can approximate $\mathbb{E}[f(X)]$ using a second order Taylor approximation of $f$ around $\mu_X:=\mathbb{E}[X]$:
\begin{align*}
\mathbb{E}[f(X)]\approx f(\mu_X)+f'(\mu_X)\mathbb{E}[X-\mu_X]+\frac{1}{2}f''(\mu_X)\mathbb{E}[X-\mu_X]^2
\end{align*}
Using such an approximation for $\mathbb{E}[b(\bm{x}_i^T(\bm{\beta}\odot\bm{\xi}_i))]$ gives $\mathbb{E}[b(\bm{x}_i^T(\bm{\beta}\odot\bm{\xi}_i))]-b(\bm{x}_i^T\bm{\beta})
\approx b''(\bm{x}_i^T\bm{\beta})\mathbb{V}[\bm{x}_i^T(\bm{\beta}\odot\bm{\xi}_i)]/2$. 
Plugging it into the penalty gives
\begin{align}
\label{eq:penalty_transformed}
\frac{1}{2}\sum_{i=1}^n \gamma_i b''(\bm{x}_i^T\bm{\beta})\mathbb{V}[\bm{x}_i^T(\bm{\beta}\odot\bm{\xi}_i)]=\frac{1}{2}\sigma^2_\mu\bm{\beta}^T\text{diag}(\bm{X}^T\bm{W}\bm{X})\bm{\beta},
\end{align}
where $\bm{X}\in\mathbb{R}^{n,d_\mu}$ is the design matrix of the mean and the diagonal weight matrix $\bm{W}$ depends on both $\bm{\beta}$ and $\bm{\alpha}$ and is given by
\begin{align*}
\bm{W}:=\text{diag}\left(\frac{\gamma_1b''(\bm{x}_1^T\bm{\beta})}{\phi/\nu_1},\ldots,\frac{\gamma_nb''(\bm{x}_n^T\bm{\beta})}{\phi/\nu_n}\right)\in\mathbb{R}^{n,n}.
\end{align*}
Since $b$ is convex, $b''>0$, and since $\gamma_i$ is nonnegative the weight matrix $\bm{W}$ is positive definite. The same holds true for $\text{diag}(\bm{X}^T\bm{W}\bm{X})$. Substituting the right-hand side of \eqref{eq:penalty_transformed} into \eqref{eq:mean_objective_transformed} shows that \eqref{eq:mean_objective_transformed2} approximates \eqref{eq:mean_objective}.

\subsection{Fisher information}
\label{app:fisher_info}

Given some log-likelihood $\ell:\Theta\rightarrow\mathbb{R}$, the Fisher information is defined by $\mathcal{I}(\theta):=\mathbb{E}_\theta[\ell'(\theta)\ell'(\theta)^T]$. Under certain regularity conditions (see Proposition 3.1 in \cite{Sha2003}) the so-called information equality
\begin{align*}
\mathcal{I}(\theta)=-\mathbb{E}_\theta[\ell''(\theta)]
\end{align*}
holds, where $\ell''(\theta)=H(\ell)$ with $H(\ell)$ is the Hessian of the log-likelihood $\ell$. Our statistical model is a product model $(\mathcal{Y}^n,\mathcal{F}^{\otimes n},(\mathbb{P}^{\otimes n}_\theta)_{\theta\in\Theta})$, which fulfills these regularity conditions. From the approximated log-likelihood
\begin{align*}
\ell(\bm{\beta},\bm{\alpha})= \frac{1}{2}\bm{z}^T\bm{\alpha}+\exp(\bm{z}^T\bm{\alpha})\frac{y\bm{x}^T\bm{\beta}-b(\bm{x}^T\bm{\beta})}{\phi/\nu}+(1-\exp(\bm{z}^T\bm{\alpha}))\frac{y\theta(y)-b(\theta(y))}{\phi/\nu}
\end{align*}
we first calculate the first derivatives with respect to $\bm{\beta}$ and $\bm{\alpha}$:
\begin{align*}
\frac{\partial\ell}{\partial\bm{\beta}}(\bm{\beta},\bm{\alpha})
&=\bm{x}^T\left\{\frac{\exp(\bm{z}^T\bm{\alpha})}{\phi/\nu}(y-b'(\bm{x}^T\bm{\beta}))\right\}\\
\frac{\partial\ell}{\partial\bm{\alpha}}(\bm{\beta},\bm{\alpha})
&=\bm{z}^T\left\{\frac{1}{2}+\frac{\exp(\bm{z}^T\bm{\alpha})}{\phi/\nu}(y\bm{x}^T\bm{\beta}-b(\bm{x}^T\bm{\beta})-(y\theta(y)-b(\theta(y))))\right\}
\end{align*}
We then proceed to compute all second derivatives:
\begin{align*}
\frac{\partial\ell}{\partial\bm{\beta}\partial\bm{\beta}^T}(\bm{\beta},\bm{\alpha})&=-\bm{x}^T\bm{x}\left\{\frac{\exp(\bm{z}^T\bm{\alpha})}{\phi/\nu}b''(\bm{x}^T\bm{\beta})\right\}\\
\frac{\partial\ell}{\partial\bm{\beta}\partial\bm{\alpha}^T}(\bm{\beta},\bm{\alpha})&=\bm{x}^T\bm{z}\left\{\frac{\exp(\bm{z}^T\bm{\alpha})}{\phi/\nu}(y-b'(\bm{x}^T\bm{\beta}))\right\}\\
\frac{\partial\ell}{\partial\bm{\alpha}\partial\bm{\alpha}^T}(\bm{\beta},\bm{\alpha})&=\bm{z}^T\bm{z}\left\{\frac{\exp(\bm{z}^T\bm{\alpha})}{\phi/\nu}(y\bm{x}^T\bm{\beta}-b(\bm{x}^T\bm{\beta})-(y\theta(y)-b(\theta(y))))\right\}\\
\frac{\partial\ell}{\partial\bm{\alpha}\partial\bm{\beta}^T}(\bm{\beta},\bm{\alpha})&=\bm{z}^T\bm{x}\left\{\frac{\exp(\bm{z}^T\bm{\alpha})}{\phi/\nu}(y-b'(\bm{x}^T\bm{\beta}))\right\}
\end{align*}
Assuming that the linear parametric assumption holds, i.e. $\mathbb{E}[Y]=b'(\bm{x}^T\bm{\beta})$ for some true parameter value $\bm{\beta}$, we obtain
\begin{align*}
\mathcal{I}(\bm{\beta},\bm{\alpha})=
\begin{bmatrix}
\mathcal{I}(\bm{\beta})&\bm{0}\\
\bm{0}&\mathcal{I}(\bm{\alpha})
\end{bmatrix}
=
\begin{bmatrix}
w_{\bm{\beta}}(\bm{x},\bm{z})\bm{x}^T\bm{x}&\bm{0}\\
\bm{0}&w_{\bm{\alpha}}(\bm{x},\bm{z})\bm{z}^T\bm{z}
\end{bmatrix},
\end{align*}
where the weights $w_{\bm{\beta}}(\bm{x},\bm{z})$ and $w_{\bm{\alpha}}(\bm{x},\bm{z})$ of the respective blocks are given by
\begin{align*}
w_{\bm{\beta}}(\bm{x},\bm{z})&:=\left\{\frac{\exp(\bm{z}^T\bm{\alpha})}{\phi/\nu}b''(\bm{x}^T\bm{\beta})\right\},\\
w_{\bm{\alpha}}(\bm{x},\bm{z})&:=\left\{\frac{\exp(\bm{z}^T\bm{\alpha})}{\phi/\nu}(b(\bm{x}^T\bm{\beta})-b'(\bm{x}^T\bm{\beta})\bm{x}^T\bm{\beta}+\mathbb{E}[Y\theta(Y)]-\mathbb{E}[b(\theta(Y))])\right\}.
\end{align*}
The term $w_{\bm{\alpha}}(\bm{x},\bm{z})$ is approximately $1$.  
To see this, observe that $w_{\bm{\alpha}}(\bm{x},\bm{z})$ is $\exp(\bm{z}^T\bm{\alpha})$ times the expected  
deviance for the exponential family used in constructing
the DEF.  Equation (2.15) in \citet{Efr1986} implies that the expected deviance is $1/\gamma=\exp(-\bm{z}^T\bm{\alpha})$ 
in a certain limiting regime, giving $w_{\bm{\alpha}}(\bm{x},\bm{z})\approx 1$. The observed Fisher information $\widehat{\mathcal{I}}(\bm{\beta})$ is an estimator of the Fisher information $\mathcal{I}(\bm{\beta})$ and it is given by
\begin{align*}
\widehat{\mathcal{I}}(\bm{\beta}):=\frac{1}{n}\sum_{i=1}^n-\frac{\partial\ell_i}{\partial\bm{\beta}\partial\bm{\beta}^T}(\bm{\beta},\bm{\alpha})=\frac{1}{n}\sum_{i=1}^nw_{\bm{\beta}}(\bm{x}_i,\bm{z}_i)\bm{x}_i^T\bm{x}_i=\frac{1}{n}\bm{X}^T\bm{W}\bm{X},
\end{align*}
showing that $\bm{\Theta}=\text{diag}(n\widehat{\mathcal{I}}(\bm{\beta}))^{1/2}$. With $w_{\bm{\alpha}}(\bm{x},\bm{z})\approx 1$ we obtain analogously
\begin{align*}
\widehat{\mathcal{I}}(\bm{\alpha}):=\frac{1}{n}\sum_{i=1}^n-\frac{\partial\ell_i}{\partial\bm{\alpha}\partial\bm{\alpha}^T}(\bm{\beta},\bm{\alpha})=\frac{1}{n}\sum_{i=1}^nw_{\bm{\alpha}}(\bm{x}_i,\bm{z}_i)\bm{z}_i^T\bm{z}_i\approx\frac{1}{n}\bm{Z}^T\bm{Z},
\end{align*}
such that $\bm{\Gamma}\approx\text{diag}(n\widehat{\mathcal{I}}(\bm{\alpha}))^{1/2}$.

\subsection{Approximations}
\label{app: approximations}

The derivations in Sections \ref{sec: Dropout regularization for the mean parameter} and \ref{subsec:dropout_mean_disp} hinge on three approximations: the approximate unity of the normalizing constant of the DEF, a second-order Taylor approximation and an argument based on the central limit theorem. These numerical approximations are widely recognized as standard tools when studying asymptotics of estimators. In our case, the use of these approximations makes our results easy to interpret and allows us to gain a deeper understanding of dropout regularization in extended GLMs. However, as explained above, these approximations might be inadequate in some settings. Here, we will discuss their limits and quality in more detail.

\subsubsection{Normalizing constant}

In the construction of the DEF, \citet{Efr1986} starts  with the random variable $Y=(1/n)\sum_{i=1}^n Z_i$ being the average of $n$ i.i.d. quantities $Z_i\sim \text{EF}(\theta)$ with densities $f_\theta$. The implied density $g_\theta$ of $Y$ is then given by
\begin{equation}
\label{eq:ef_efron}
    g_\theta(y)=\prod_{i=1}^nf_\theta(y)=h(y)^n\exp(n\theta y-nb(\theta)),
\end{equation}
which is simply the joint density of $Z_1,\ldots,Z_n$. It is shown that the normalizing constant of the DEF constructed from \eqref{eq:ef_efron} fulfills $C(\gamma,\theta)=1+\mathcal{O}(n^{-1})$. In other words, given a sample of size $n$ the error induced by the assumption $C(\gamma,\theta)\approx1$ is of order $\mathcal{O}(n^{-1})$, making it negligible in most applications.

\subsubsection{Taylor approximation} 

\citet{WagWanLia2013} provide an empirical analysis of the accuracy of the second-order Taylor approximation in case of logistic regression and Gaussian dropout noise. They compare the true penalty and its quadratic approximation for different levels of the mean $p$ and across a range of standard deviations $\sigma$. The approximation appears to be very accurate, with slight overestimation for $p\approx0.5$ and slight underestimation for very confident predictions.

More generally, the observation-wise error resulting from the quadratic Taylor approximation described in Section \ref{app:dropout_mean} is given by the expected remainder
\begin{equation}
    \label{eq:expected_remainder}
\mathbb{E}[R(\bm{x}^T(\bm{\beta}\odot\bm{\xi}))] = \mathbb{E}\left[\sum_{k=3}^\infty \frac{b^{(k)}(\bm{x}^T\bm{\beta})}{k!}(\bm{x}^T(\bm{\beta}\odot\bm{\xi})-\bm{x}^T\bm{\beta})^k\right],
\end{equation}
which shows that the error depends on the particular distribution of the dependent variable through the derivatives $b^{(k)}$ of the partition function $b$ as well as on the distribution of the dropout noise $\bm{\xi}$ through the standardized moments $\mathbb{E}[(\bm{x}^T(\bm{\beta}\odot\bm{\xi})-\bm{x}^T\bm{\beta})^k]$ and the data $\bm{x}$ itself.

First of all, we have $b(x)=x^2/2$ in case of Gaussian data, thus the approximation will be exact since $b^{(k)}=0$ for $k\geqslant3$. For Poisson data $b(x)=\exp(x)$ together with $\exp(x):=\sum_{k=0}^\infty x^k/k!$ yields
\begin{align*}
\mathbb{E}[R(\bm{x}^T(\bm{\beta}\odot\bm{\xi}))] &= \exp(\bm{x}^T\bm{\beta})\mathbb{E}\left[\sum_{k=3}^\infty \frac{(\bm{x}^T(\bm{\beta}\odot\bm{\xi})-\bm{x}^T\bm{\beta})^k}{k!}\right].
\end{align*}
Consider the case of Gaussian dropout noise $\bm{\xi}\sim\mathcal{N}(\bm{1}_d,\sigma^2\mathbb{I}_d)$, then it holds that
\begin{align*}
    \bm{x}^T(\bm{\beta}\odot\bm{\xi})-\bm{x}^T\bm{\beta}\sim\mathcal{N}(0,\sigma^2 s^2) \text{ with } \sigma^2 s^2:= \sigma^2\sum_{j=1}^{d_\mu} (x_j\beta_j)^2.
\end{align*}
It is well known that for the centralized moments of a Gaussian distributions the odd moments vanish and the even moments compute to $(\sigma^2s^2)^n(n-1)!!$ where $n!!$ denotes the double factorial. Since the non-vanishing part is an infinite sum of non-negative random variables by the monotone convergence theorem it holds
\begin{align}
\notag
\mathbb{E}[R(\bm{x}^T(\bm{\beta}\odot\bm{\xi}))] 
&= \exp(\bm{x}^T\bm{\beta})\sum_{k=2}^\infty \frac{(\sigma^2s^2)^{2k} (2k-1)!!}{2k!} \\
\notag
&= \exp(\bm{x}^T\bm{\beta})\sum_{k=2}^\infty \frac{(\sigma^2s^2)^{2k}}{2k!!} \\
\label{eq:rem_poisson}
&= \exp(\bm{x}^T\bm{\beta})\left\{\exp\bigg(\frac{(\sigma^2s^2)^2}{2}\bigg)-\frac{(\sigma^2s^2)^2}{2}-1\right\},
\end{align}
where we used the relation $n!=n!!(n-1)!!$ and the fact that this infinite sum has an explicit limit that can be derived from $\sum_{k=0}^\infty x^{2k}/2k!!= \exp(x^2/2)$.

Now consider the case of Bernoulli dropout noise. The arbitrarily weighted sum of i.i.d. Bernoulli random variables does not obey an easy distributional equality as in the classical case of a simple sum. A detailed analysis could be the subject of future work. However, since generally $d$ is relatively large we can assume that $\bm{x}^T(\bm{\beta}\odot\bm{\xi})$ is approximately Gaussian. Therefore the Gaussian case gives a rough indication.

Lastly, let us address the case of Binomial data. Then the partition function is $b(x)=N\log(1+\exp(x))$, hence infinitely often continuously differentiable. In particular all derivatives can be bounded by $1$ (even smaller values for many derivatives are possible). Therefore the remainder admits an expansion as in \eqref{eq:expected_remainder} and we can write in the case of Gaussian dropout with the exact same arguments as above
\begin{align}
\label{eq:rem_binomial}
\vert\mathbb{E}[R(\bm{x}^T(\bm{\beta}\odot\bm{\xi}))]\vert &\leq \sum_{k=2}^\infty \frac{(\sigma^2s^2)^{2k}}{2k!!}=\left\{\exp\bigg(\frac{(\sigma^2s^2)^2}{2}\bigg)-\frac{(\sigma^2s^2)^2}{2}-1\right\}.
\end{align}
The considerations w.r.t. to the Bernoulli dropout noise are analogous to the Poisson case.

In sum, both \eqref{eq:rem_poisson} and \eqref{eq:rem_binomial} show that the approximation error will be small whenever $\sigma^2$ is small and in case of Poisson data, if the mean $\exp(\bm{x}^T\bm{\beta})$ is small. This however was already obvious from \eqref{eq:expected_remainder}, which shows that $\sigma^2=0$ implies $\mathbb{E}[R(\bm{x}^T(\bm{\beta}\odot\bm{\xi}))] =0$. But more importantly, the error will be small if
\begin{align*}
    s^2=\sum_{j=1}^{d_\mu} (x_j\beta_j)^2
\end{align*}
is small. And this will be the case when there are rare but important features, which is the most fundamental characteristic for dropout regularization in a regression context.

\subsubsection{Central limit theorem.} The normality assumption leading to \eqref{eq: log_gamma_normal_approx} is of course exact in the case of Gaussian dropout noise. As noted in the previous paragraph, the case of Bernoulli dropout noise is not trivial and could be the subject of future research.

\subsection{Details on the SGD algorithm} \label{app: algorithm}

We use SGD to estimate the parameters $\bm{\beta}$ and $\bm{\alpha}$ which parameterize the linear predictors $\theta(x)=\bm{B}_\mu(x)^T\bm{\beta}$ and $\log(\gamma(x))=\bm{B}_\gamma(x)^T\bm{\alpha}$, where $\bm{B}_\mu(x)$ and $\bm{B}_\gamma(x)$ are vectors of all evaluated B-spline basis functions at $x\in[a,b]$. For given hyperparameters $\sigma^2_\mu$ and $\sigma^2_\gamma$ and after setting initial parameter values $\bm{\beta}^{(0)}$ and $\bm{\alpha}^{(0)}$ the estimation procedure starts performing SGD-updates on both parameter vectors. At every iteration $i$, for given estimates $\bm{\beta}^{(i-1)}$ and $\bm{\alpha}^{(i-1)}$ it performs the updates
\begin{align*}
\bm{\beta}^{(i)}&=\bm{\beta}^{(i-1)}+\eta^{(i)}_\mu s^{(i)}(\bm{\beta}^{(i-1)}),\\
\bm{\alpha}^{(i)}&=\bm{\alpha}^{(i-1)}+\eta^{(i)}_\gamma s^{(i)}(\bm{\alpha}^{(i-1)}).
\end{align*}
The scores $s^{(i)}(\bm{\beta}^{(i-1)})$ and $s^{(i)}(\bm{\alpha}^{(i-1)})$ are estimates based on a batch sample, which is perturbed by Gaussian or Bernoulli dropout noise. The batch sample is obtained by sampling $b=30$ indices uniformly from the index set $\{1,\ldots,n\}$. The learning rates $\eta^{(i)}_\mu$ and $\eta^{(i)}_\gamma$ are determined by \texttt{ADADELTA} with default values suggested in \citep{Zei2012}. The procedure terminates if a maximum number of iterations has been performed or if the log-likelihood reaches a stationary point.

We perform random search $k$-fold CV in order to select $\sigma_\mu$ and $\sigma_\gamma$ from the hyperparameter space $\mathcal{H}\subseteq\mathbb{R}_+^2$ optimally.  That is, after creating $k\in\mathbb{N}$ folds from the data $\{\bm{y},\bm{B}_\mu,\bm{B}_\gamma\}$ we draw $s\in\mathbb{N}$ samples $\{(\sigma_{\bm{\xi},1},\sigma_{\bm{\zeta},1}),\ldots,(\sigma_{\bm{\xi},s},\sigma_{\bm{\zeta},s})\}$ from the uniform distribution over $\mathcal{H}$. Then, for each tuple $(\sigma_{\bm{\xi},j},\sigma_{\bm{\zeta},j})$ and $j=1,\ldots,s$ we estimate the model $k$ times. Each of these $k$ estimates will be based on all but one fold. The left-out fold is used to evaluate each estimated model by calculating the log-likelihood. The average over the $k$ log-likelihood values corresponding to the $j$-th sample is given by $\ell_j$. The optimal smoothing parameters $\sigma_\mu^*$ and $\sigma_\gamma^*$ are set according to
\begin{align*}
\label{eq:cv_optimum}
(\sigma_\mu^*,\sigma_\gamma^*)\,\gets\,(\sigma_{\bm{\xi},j^*},\sigma_{\bm{\zeta},j^*}),
\end{align*}
where the index $j^*$ is defined as $\label{eq:cv_optimum_index}
j^*=\argmax_{j\in\{1,\ldots,s\}}\{\ell_1,\ldots,\ell_s\}$.

The code is publicly available through Github\footnote{\url{https://github.com/luetkeschwienhorst/dropoutinextendedGLMs}}. It relies on parallel computing and all computations were performed on a GPU with 192 GB memory and 12 physical cores at 3.0 GHz. It also contains the code from \citet{GijProCla2010} obtained via an URL provided in their publication.

\subsection{Details on the simulation design}\label{app: simulation design}

\subsubsection*{Testfunctions}

As mentioned before, dropout favors rare but important features. The true mean and dispersion function from which we simulate data should therefore allow for this characteristic, in view of the chosen functional basis. We use B-splines with a relatively large number of knots, such that the basis functions are only supported on small compact sets of the domain. If the true function to be approximated with these basis functions is mostly flat, but spikes on small subsets of the domain, then we have correctly chosen a basis of rare but important features. All testfunctions were constructed with this idea in mind, as can be seen in Figure \ref{fig:testfunctions}. This includes the intentional choice $g_2$ in order to verify if dropout regularization deteriorates in this case.

The testfunctions for the mean are
\begin{align}
f_1&:[0,1]\rightarrow\mathbb{R},\quad x\mapsto 2\sin(4\pi x)\phi_{0.5,0.05^2}(x),\\
f_2&:[0,1]\rightarrow\mathbb{R},\quad x\mapsto 40+10\sin(4\pi x)\phi_{0.5,0.05^2}(x),
\end{align}
whereas $f_1$ was used to simulate normal data and $f_2$ served as mean function in the Poisson and binomial case. We will fix the number of trials to $N=70$ when using $f_3$ to model double binomial data. The function $\phi_{\mu,\sigma^2}$ is the PDF of the normal distribution $\mathcal{N}(\mu,\sigma^2)$. For each distribution, we combined the mean function with one of the three dispersion functions
\begin{align}
g_1&:[0,1]\rightarrow\mathbb{R},\quad x\mapsto \exp\{-0.08\phi_{0.4,0.04^2}(x)-0.08\phi_{0.7,0.02^2}(x)\},\\
g_2&:[0,1]\rightarrow\mathbb{R},\quad x\mapsto \exp\{-0.08\phi_{0.4,0.03^2}(x)+0.08\phi_{0.7,0.03^2}(x)\},\\
g_3&:[0,1]\rightarrow\mathbb{R},\quad x\mapsto \exp\left\{\frac{2}{0.15}\phi_{0,1}\left(\frac{x-0.8}{0.15}\right)\Phi_{0,1}\left(-4\left(\frac{x-0.8}{0.15}\right)\right)\right\}.
\end{align}

\subsubsection*{Simulation design}

For a fixed distribution $\mathbb{P}\in\{\mathcal{N},\text{DPoi},\text{DBin}\}$, mean function $f_i$ with $i\in\{1,2\}$, dispersion function $g_j$ with $j\in\{1,2,3\}$ and for a sample size of $n=250,500,1000$, the simulation procedure consists of the following steps:
\begin{enumerate}
\item Simulate $R+1$ with $R=100$ data sets $\mathcal{D}(\mathbb{P},i,j,n)[m]=\{(y_k,x_k)\}_{k=1}^n$ for $m=1,\ldots,R+1$, i.e. we denote by $\mathcal{D}(\mathbb{P},i,j,n)[m]$ the $m$-th data set of sample size $n$ which was simulated from the distribution $\mathbb{P}$ with mean $f_i$, dispersion $g_j$. The values $x_k$ were sampled from $\mathcal{U}([0,1])$. Each dependent variable $y_k$ was then sampled from the respective $\text{DEF}(\gamma_k,\mu_k,\phi)$ with $\gamma_k=g_j(x_k)$ and $\mu_k=f_i(x_k)$. We fixed $\sigma^2=0.8^2$ in the normal case.
\item Perform $k$-fold likelihood cross-validation for $k=5$ on the first replicate $\mathcal{D}(\mathbb{P},i,j,n)[1]$ for each regularization method, where the number of drawn samples from the uniform distribution over $\mathcal{H}$ is fixed to $s=500$:
\label{item:k_fold_cv}
\begin{enumerate}
\item For Bernoulli dropout fix $\mathcal{H}=\mathcal{I}_\mu\times\mathcal{I}_\gamma=[0,1]^2$.
\item For Gaussian dropout fix $\mathcal{H}=\mathcal{I}_\mu\times\mathcal{I}_\gamma=[0,3]\times[0,6]$.
\item For PMLE fix $\mathcal{H}=\mathcal{I}_\mu\times\mathcal{I}_\gamma=[0,15000]^2$.
\end{enumerate}
\item Use the optimal smoothing parameters for each method obtained in (\ref{item:k_fold_cv}) to estimate for the remaining $R$ data sets $\mathcal{D}(\mathbb{P},i,j,n)[m]$ with $m=2,\ldots,R+1$ the corresponding models
$\mathcal{M}(\mathbb{P},i,j,n,\mathcal{R})[m]$. By $\mathcal{M}(\mathbb{P},i,j,n,\mathcal{R})[m]$ we denote the model estimate obtained from data set $\mathcal{D}(\mathbb{P},i,j,n)[m]$ and regularization method $\mathcal{R}\in\{\texttt{bernoulli},\texttt{gaussian},\texttt{pmle}\}$. We will simply write $\mathcal{M}(\mathbb{P},i,j,n,\mathcal{R})$ in order to refer to all estimates. Out of convenience, we will sometimes drop $n$ and $R$ and only write $\mathcal{M}(\mathbb{P},i,j)$ when we want to refer to all models which have $\mathbb{P}$, $i$ and $j$ in common, but vary across $n$ and $\mathcal{R}$.
\item Let $\widehat{\pi}^{(m)}$ be the mean or dispersion estimate obtained from $\mathcal{D}(\mathbb{P},i,j,n)[m]$ and $\pi$ its ground truth counterpart. To evaluate the performance of the three competing regularization methods we compute the RMSE
\begin{align}
\label{eq:rmse}
\text{RMSE}(\widehat{\pi}^{(m)},\pi):=\left(\sum_{x\in\Pi}\left(\widehat{\pi}^{(m)}(x)-\pi(x)\right)^2\right)^{1/2}
\end{align}
for all $m=2,\ldots,R+1$ and an equidistant grid $\Pi:=[0,1/\kappa,2/\kappa,\ldots,1]$ of length $\kappa+1$ with $\kappa=500$.
\end{enumerate}
\noindent
We employ natural cubic B-splines bases based on $m_\mu=30$ equidistant knots in the mean and $m_\gamma=20$ equidistant knots in the dispersion model.

\newpage
\subsection{Additional plots}
\label{app:additional_plots}

\begin{figure}[htb]
\centering
\begin{subfigure}{0.24\textwidth}
\includegraphics[width=\textwidth]{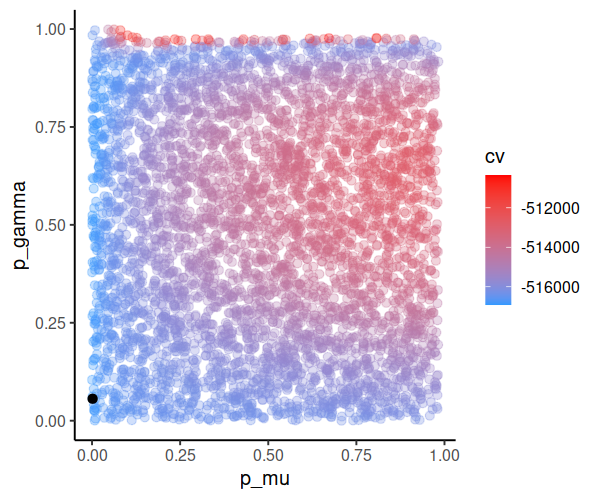}
\includegraphics[width=\textwidth]{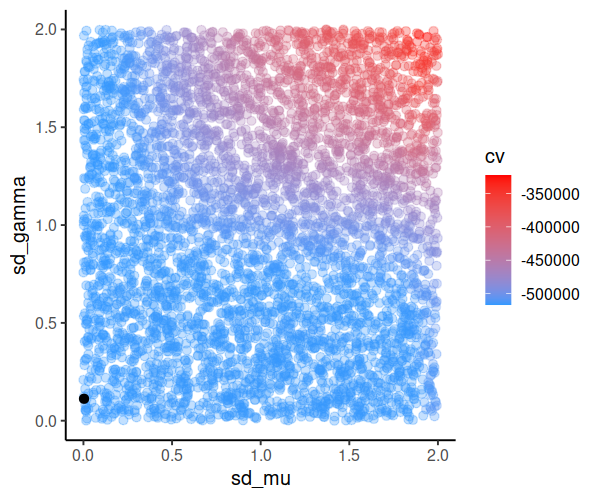}
\includegraphics[width=\textwidth]{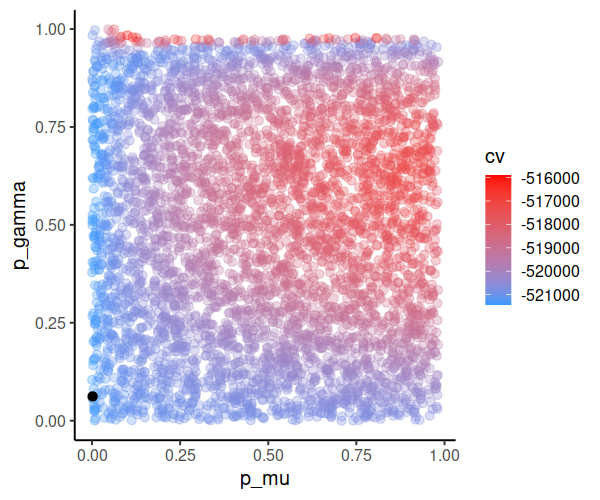}
\includegraphics[width=\textwidth]{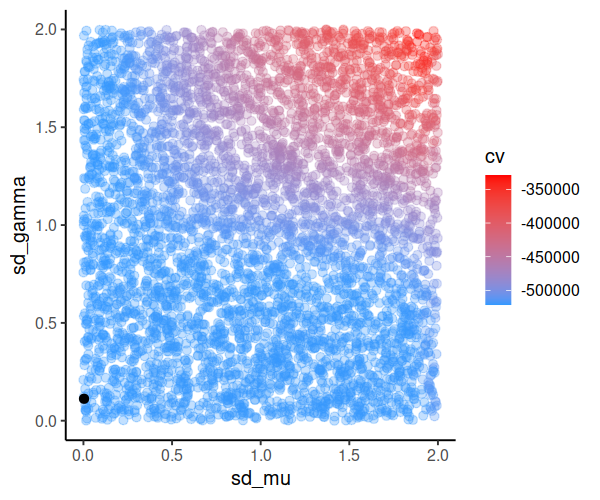}
\caption{West}
\vspace{2mm}
\end{subfigure}
\begin{subfigure}{0.24\textwidth}
\includegraphics[width=\textwidth]{west_in_scatter_ber}
\includegraphics[width=\textwidth]{west_in_scatter_norm}
\includegraphics[width=\textwidth]{west_out_scatter_ber}
\includegraphics[width=\textwidth]{west_out_scatter_norm}
\caption{South}
\vspace{2mm}
\end{subfigure}
\begin{subfigure}{0.24\textwidth}
\includegraphics[width=\textwidth]{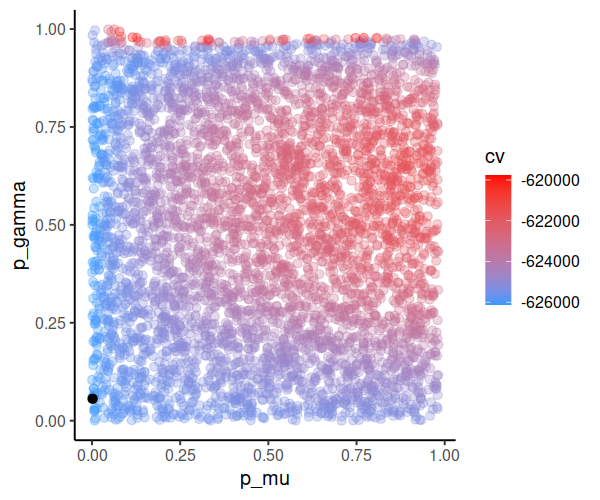}
\includegraphics[width=\textwidth]{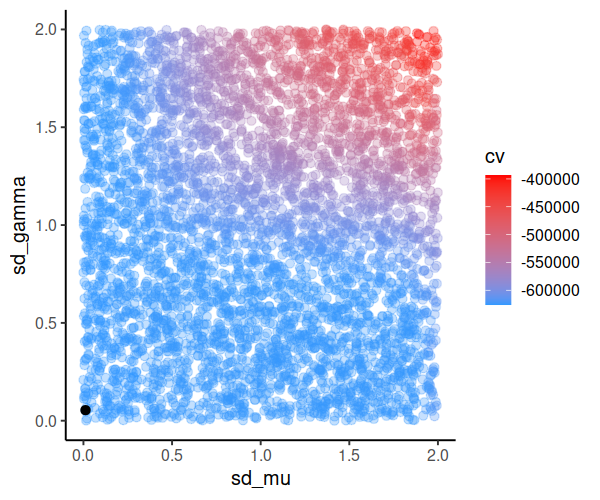}
\includegraphics[width=\textwidth]{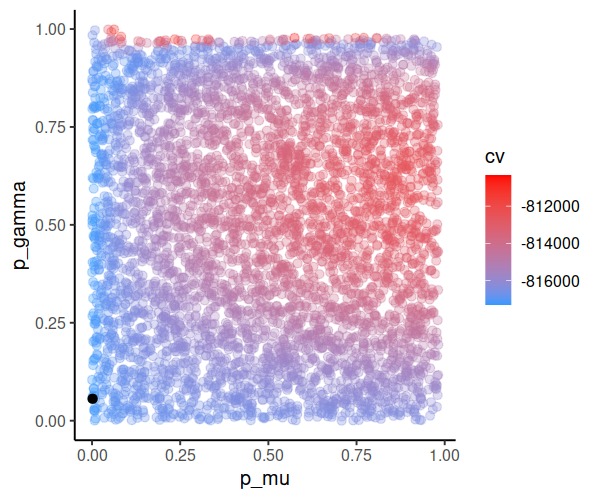}
\includegraphics[width=\textwidth]{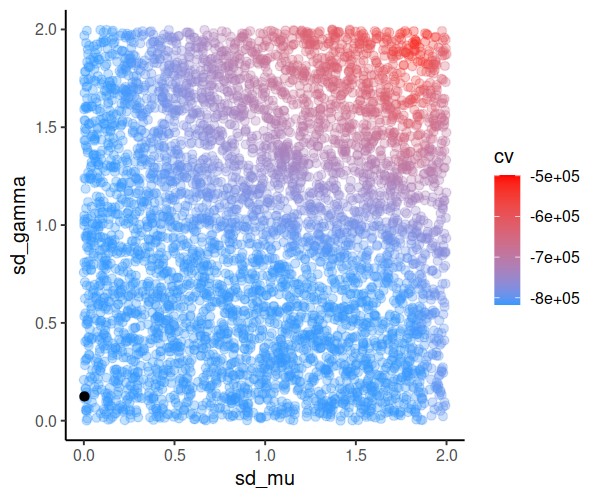}
\caption{East}
\vspace{2mm}
\end{subfigure}
\begin{subfigure}{0.24\textwidth}
\includegraphics[width=\textwidth]{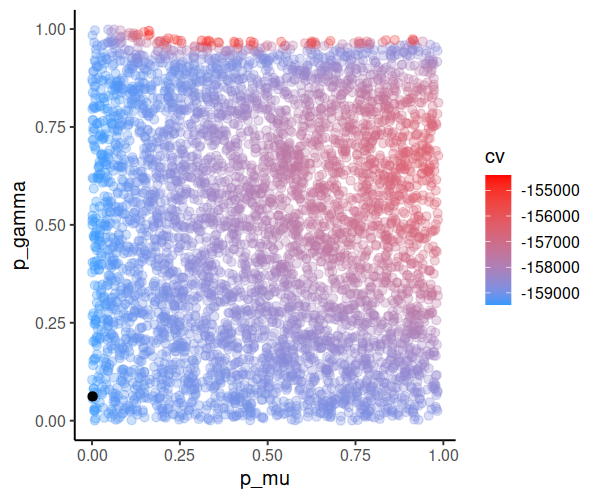}
\includegraphics[width=\textwidth]{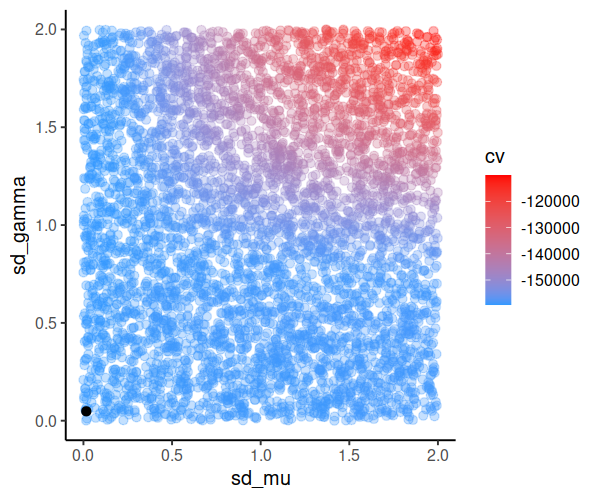}
\includegraphics[width=\textwidth]{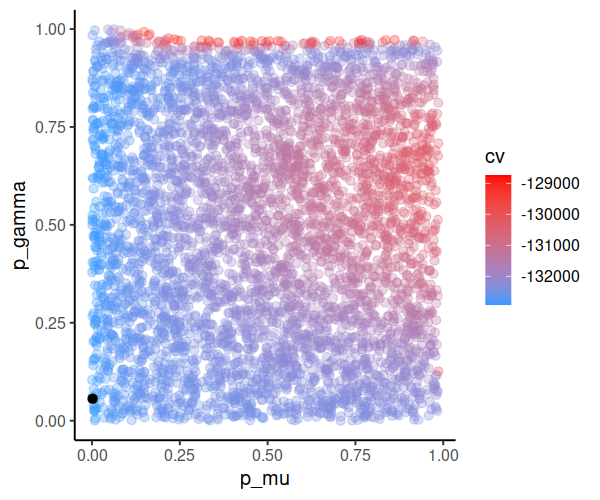}
\includegraphics[width=\textwidth]{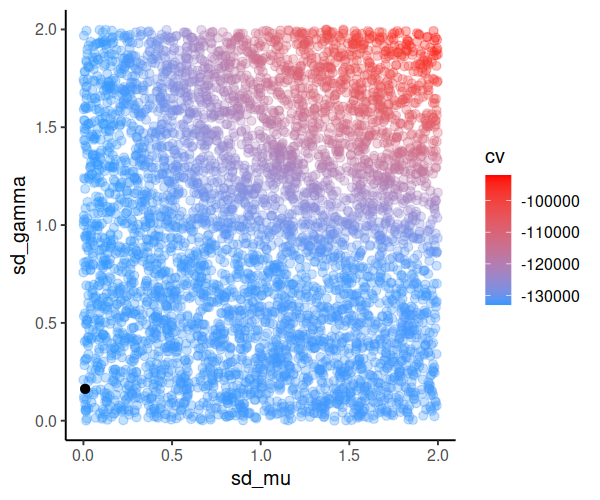}
\caption{North}
\vspace{2mm}
\end{subfigure}
\caption{Cross-validation results for the traffic detection data of the four sensors in the (a) west, (b) south, (c) east and (d) north of the Berlin city center. The top two rows depict the results for Bernoulli and Gaussian dropout of the inbound traffic respectively. The bottom two rows show the results corresponding to the outbound traffic. The optimal values are marked as a black dot in the plots. Note that 200 samples with very large values (including NaNs) were removed in case of Bernoulli dropout, in order to make the plots more informative.}
\label{fig:cv_traffic_data}
\end{figure}

\begin{figure}[ht]
\centering
\begin{subfigure}{\textwidth}
\includegraphics[width=0.29\textwidth]{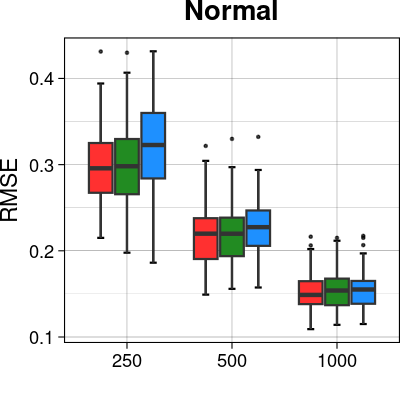}
\includegraphics[width=0.29\textwidth]{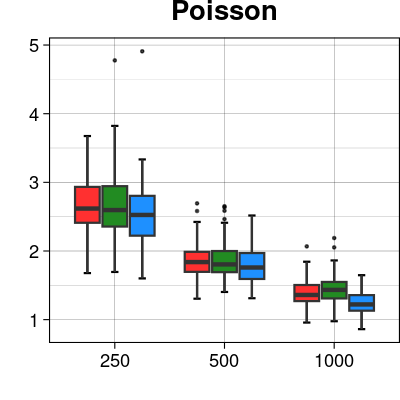}
\includegraphics[width=0.4\textwidth]{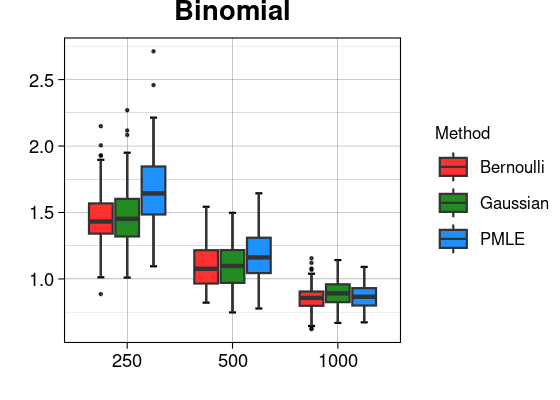}
\end{subfigure}
\begin{subfigure}{\textwidth}
\includegraphics[width=0.29\textwidth]{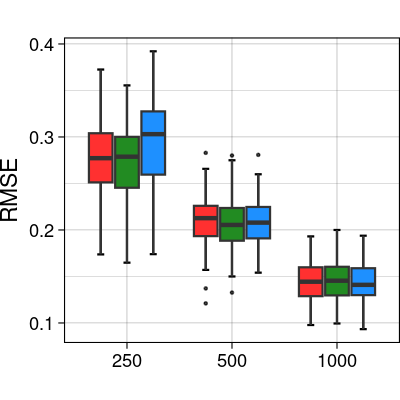}
\includegraphics[width=0.29\textwidth]{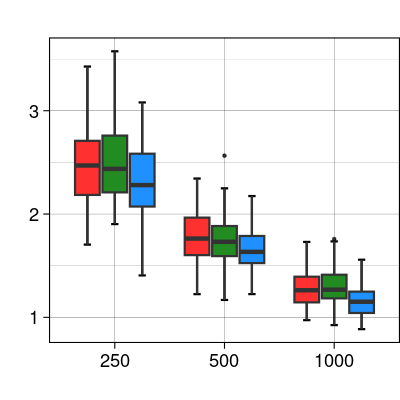}
\includegraphics[width=0.29\textwidth]{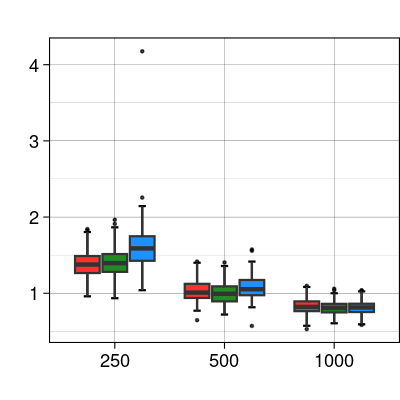}
\end{subfigure}
\begin{subfigure}{\textwidth}
\includegraphics[width=0.29\textwidth]{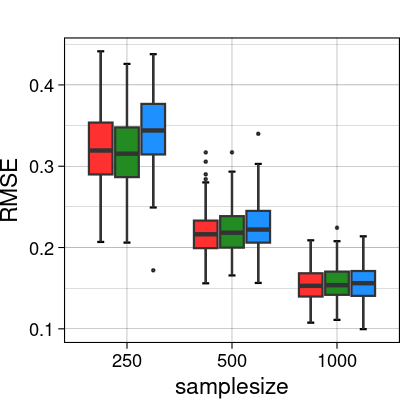}
\includegraphics[width=0.29\textwidth]{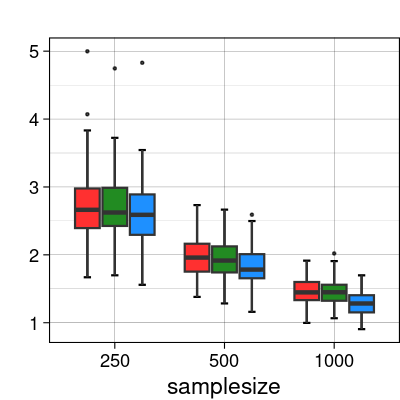}
\includegraphics[width=0.29\textwidth]{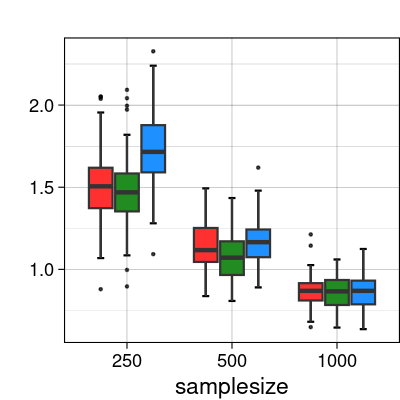}
\end{subfigure}
\caption{Boxplots of RMSEs for mean estimates of normal data (left column), Poisson data (middle column) and binomial data (right column) across Scenarios 1 (top row), 2 (middle row) and 3 (bottom row).}
\label{fig:boxplots_mean}
\end{figure}

\begin{figure}[ht]
\centering
\begin{subfigure}{\textwidth}
\includegraphics[width=\textwidth]{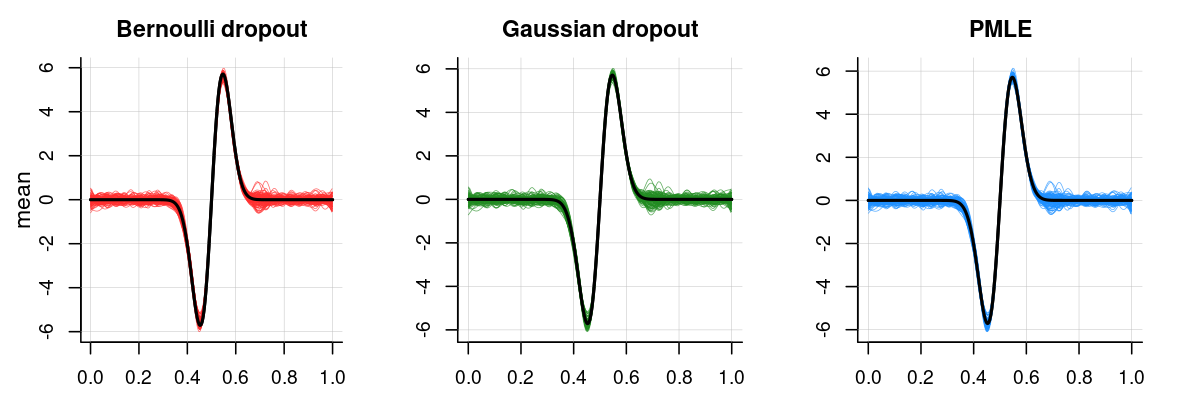}
\vspace{-5mm}
\end{subfigure}
\begin{subfigure}{\textwidth}
\includegraphics[width=\textwidth]{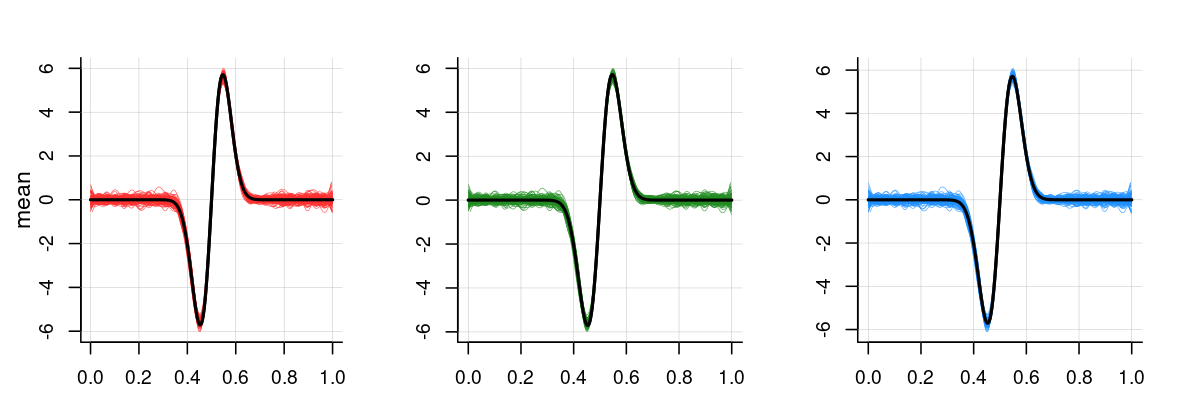}
\vspace{-5mm}
\end{subfigure}
\begin{subfigure}{\textwidth}
\includegraphics[width=\textwidth]{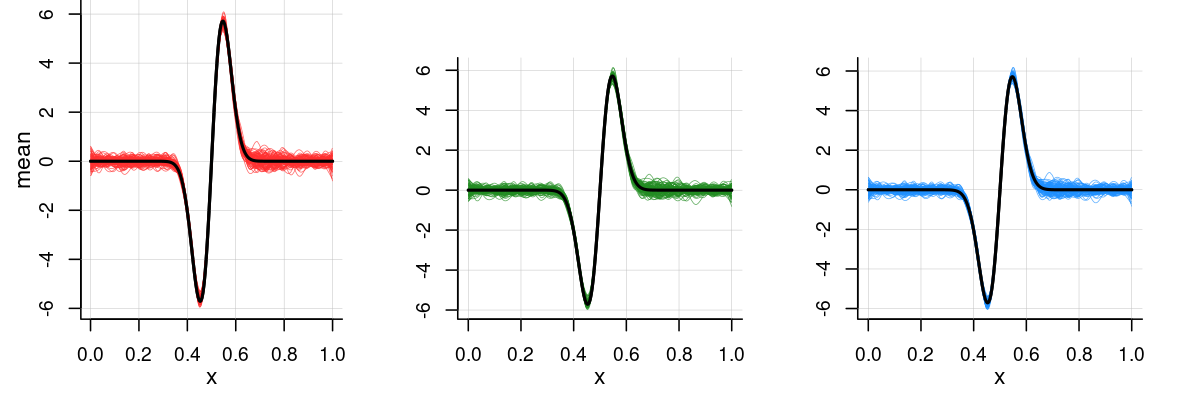}
\vspace{-4mm}
\end{subfigure}
\caption{Estimated mean effects in the normal model for Bernoulli dropout (left), Gaussian dropout (middle) and PMLE (right) in Scenario 1 (upper row), Scenario 2 (middle row) and Scenario 3 (bottom row) for $R=100$ replicates of size $n=1000$. The true effects are given by the black lines.}
\label{fig:effects_normal_mean}
\end{figure}

\begin{figure}[ht]
\centering
\begin{subfigure}{\textwidth}
\includegraphics[width=\textwidth]{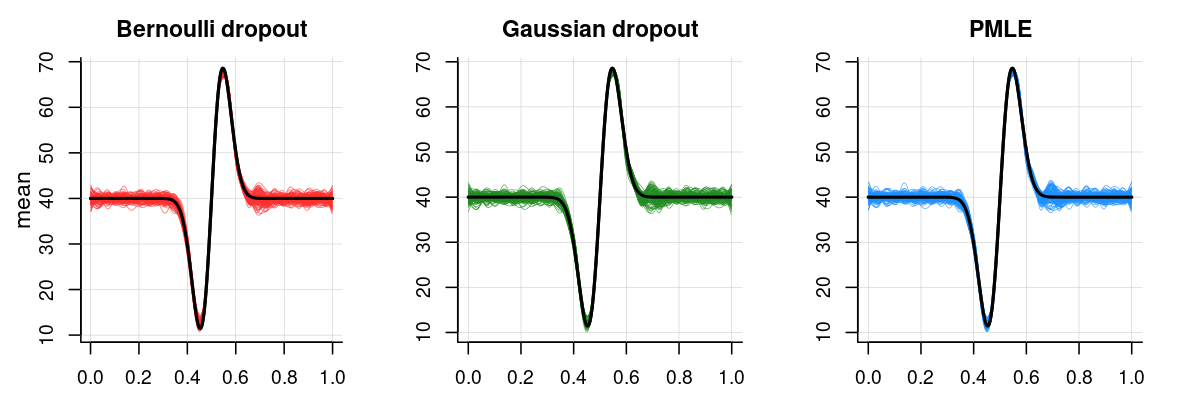}
\vspace{-5mm}
\end{subfigure}
\begin{subfigure}{\textwidth}
\includegraphics[width=\textwidth]{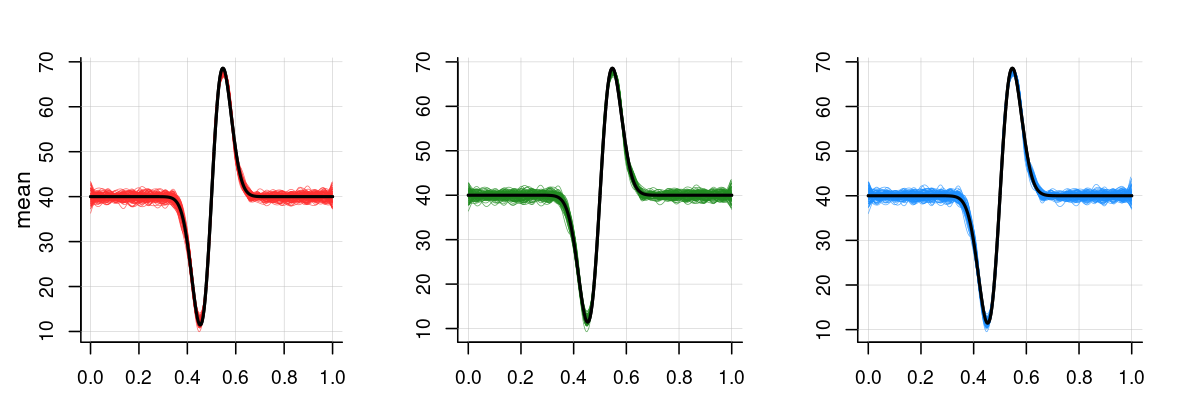}
\vspace{-5mm}
\end{subfigure}
\begin{subfigure}{\textwidth}
\includegraphics[width=\textwidth]{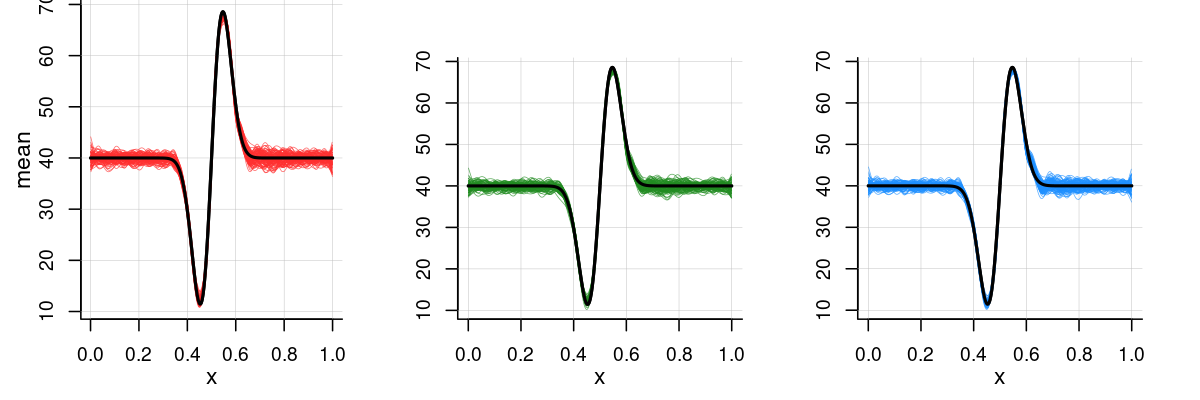}
\vspace{-4mm}
\end{subfigure}
\caption{Estimated mean effects in the binomial model for Bernoulli dropout (left), Gaussian dropout (middle) and PMLE (right) in Scenario 1 (upper row), Scenario 2 (middle row) and Scenario 3 (bottom row) for  $R=100$ replicates of size $n=1000$. The true effects are given by the black lines.}
\label{fig:effects_binomial_mean}
\end{figure}

\begin{figure}[ht]
\centering
\begin{subfigure}{\textwidth}
\includegraphics[width=\textwidth]{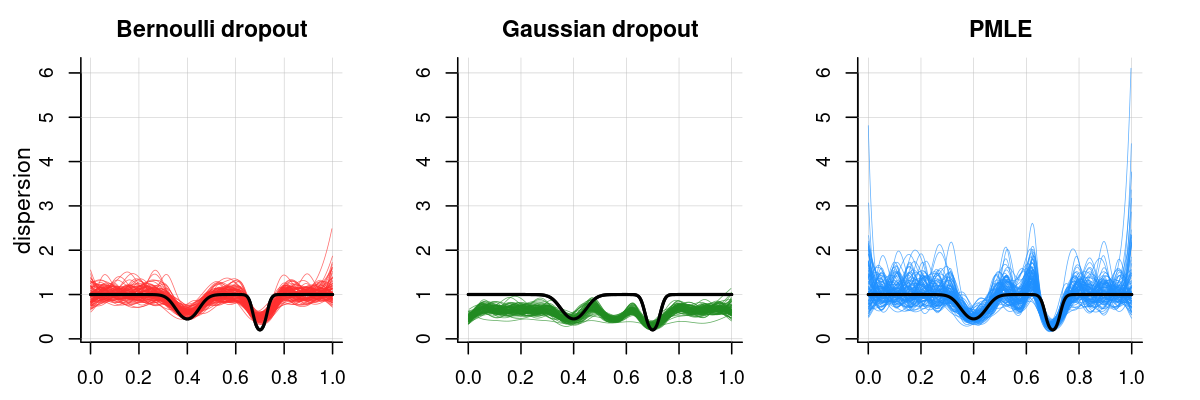}
\vspace{-5mm}
\end{subfigure}
\begin{subfigure}{\textwidth}
\includegraphics[width=\textwidth]{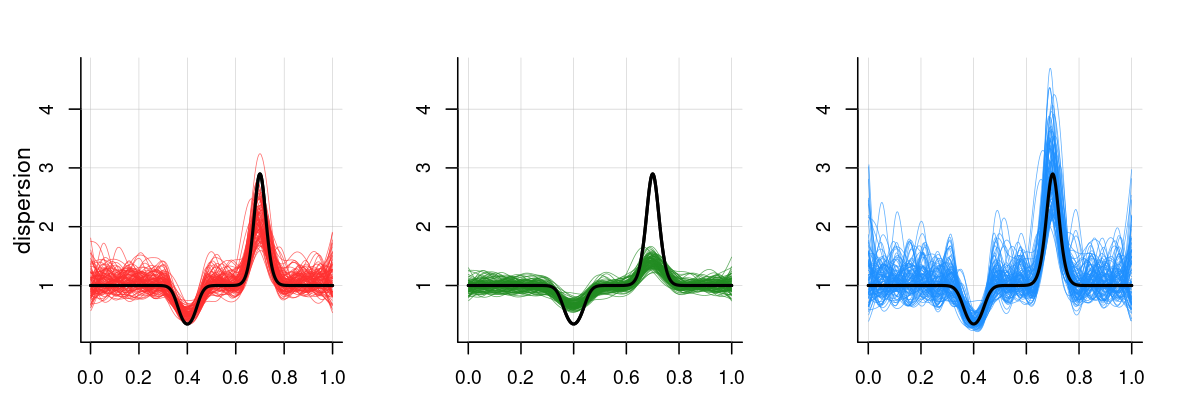}
\vspace{-5mm}
\end{subfigure}
\begin{subfigure}{\textwidth}
\includegraphics[width=\textwidth]{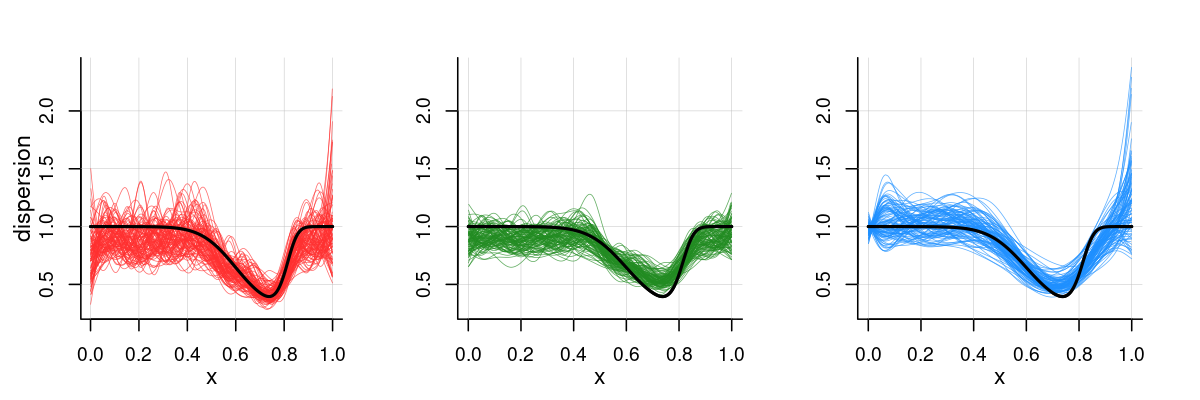}
\vspace{-4mm}
\end{subfigure}
\caption{Estimated dispersion effects in the Poisson model for Bernoulli dropout (left), Gaussian dropout (middle) and PMLE (right) in Scenario 1 (upper row), Scenario 2 (middle row) and Scenario 3 (bottom row) for  $R=100$ replicates of size $n=1000$. The true effects are given by the black lines.}
\label{fig:effects_poisson_disp}
\end{figure}

\begin{figure}[ht]
\centering
\begin{subfigure}{\textwidth}
\includegraphics[width=\textwidth]{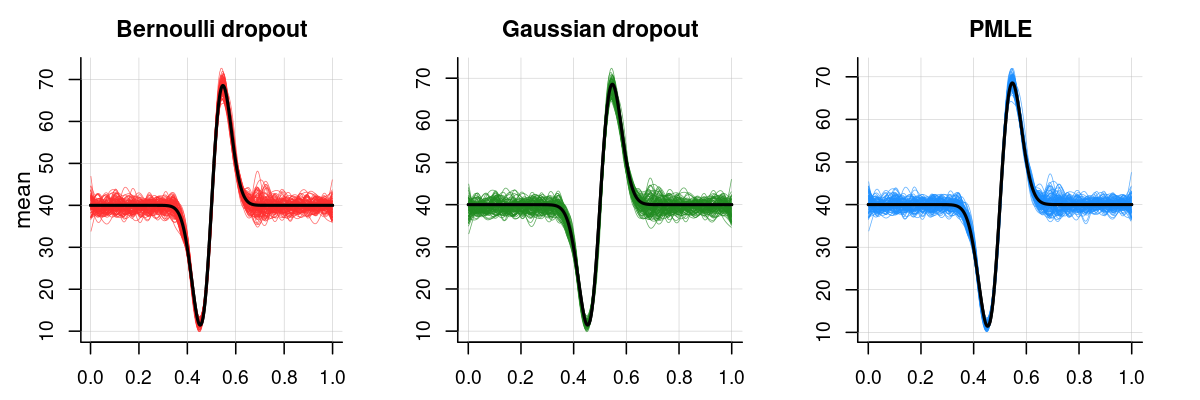}
\vspace{-5mm}
\end{subfigure}
\begin{subfigure}{\textwidth}
\includegraphics[width=\textwidth]{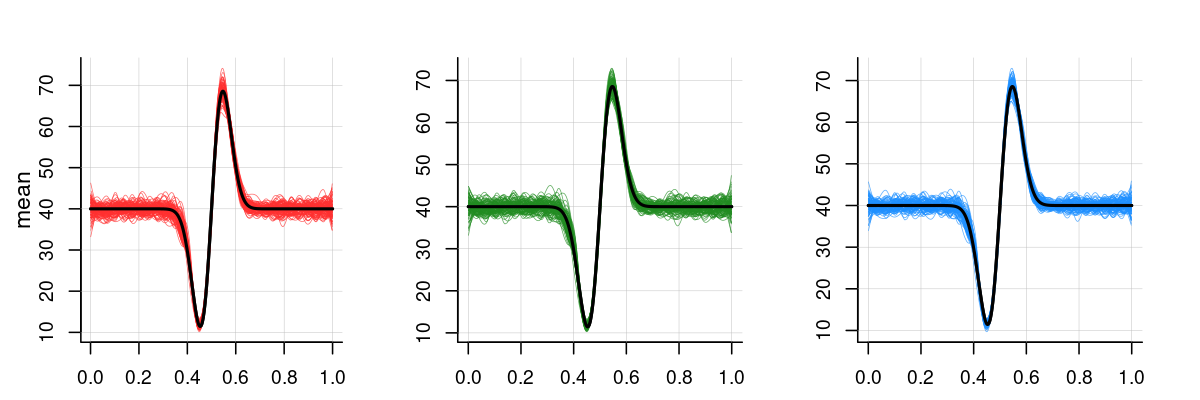}
\vspace{-5mm}
\end{subfigure}
\begin{subfigure}{\textwidth}
\includegraphics[width=\textwidth]{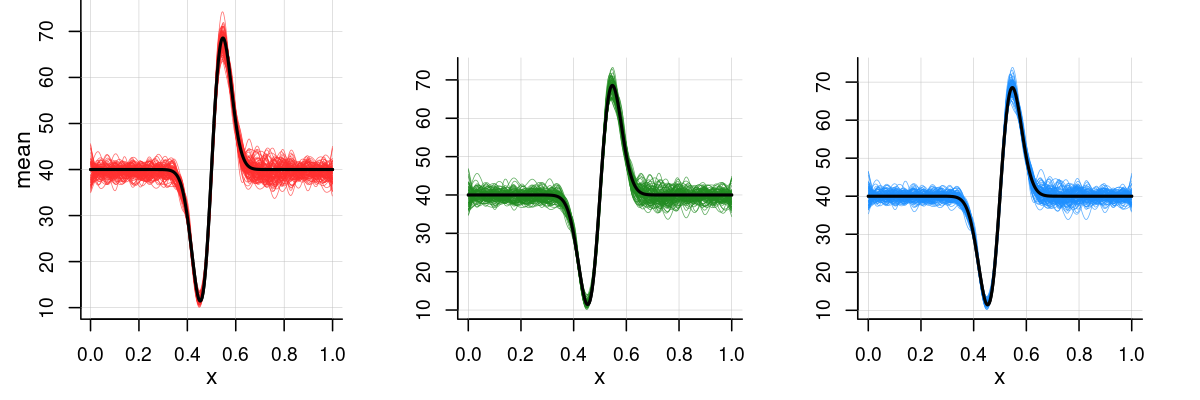}
\vspace{-4mm}
\end{subfigure}
\caption{Estimated mean effects in the Poisson model for Bernoulli dropout (left), Gaussian dropout (middle) and PMLE (right) in Scenario 1 (upper row), Scenario 2 (middle row) and Scenario 3 (bottom row) for  $R=100$ replicates of size $n=1000$. The true effects are given by the black lines.}
\label{fig:effects_poisson_mean}
\end{figure}

\end{document}